\documentclass[journal]{IEEEtran}
\ifCLASSINFOpdf
\else
\fi
\usepackage{graphicx}
\usepackage{microtype}
\usepackage{subfig}
\usepackage{booktabs} 
\usepackage{soul}
\usepackage{hyperref}
\usepackage{color,soul}
\usepackage{ctable}
\usepackage[bottom]{footmisc}
\usepackage{dirtytalk}

\begin{document}
%
\title{Weak Supervision for Generating Pixel--Level Annotations in Scene Text Segmentation}
%
%
%

\author{Simone Bonechi, Paolo Andreini, Monica Bianchini, Franco Scarselli
\thanks{The authors are with the Department
of Information Engineering and Mathematics, University of Siena, Siena, 53034 Italy e-mail: simone.bonechi@unisi.it}
\thanks{S. Bonechi and P. Andreini contribute equally to the work.}}
\maketitle

\begin{abstract}
Providing pixel--level supervisions for scene text segmentation is inherently difficult and costly, so that only few small datasets are available for this task. To face the scarcity of training data, previous approaches based on Convolutional Neural Networks (CNNs) rely on the use of a synthetic dataset for pre--training. 
However, synthetic data cannot reproduce the complexity and variability of natural images. In this work, we propose to use a weakly supervised learning approach to reduce the domain--shift between synthetic and real data. Leveraging the bounding--box supervision of the COCO--Text and the MLT datasets, we generate weak pixel--level supervisions of real images. 
In particular, the COCO--Text--Segmentation (COCO\_TS) and the MLT--Segmentation (MLT\_S) datasets are created and released.
These two datasets are used to train a CNN, the Segmentation Multiscale Attention Network (SMANet), which is specifically designed to face some peculiarities of the scene text segmentation task.
The SMANet is trained end--to--end on the proposed datasets, and the experiments show that COCO\_TS and MLT\_S are a valid alternative to synthetic images, allowing to use only a fraction of the training samples and improving significantly the performances.
\end{abstract}


%
\IEEEpeerreviewmaketitle

\section{Introduction}
\begin{figure*}[!h]
\begin{center}
\centerline{\includegraphics[height=8cm]{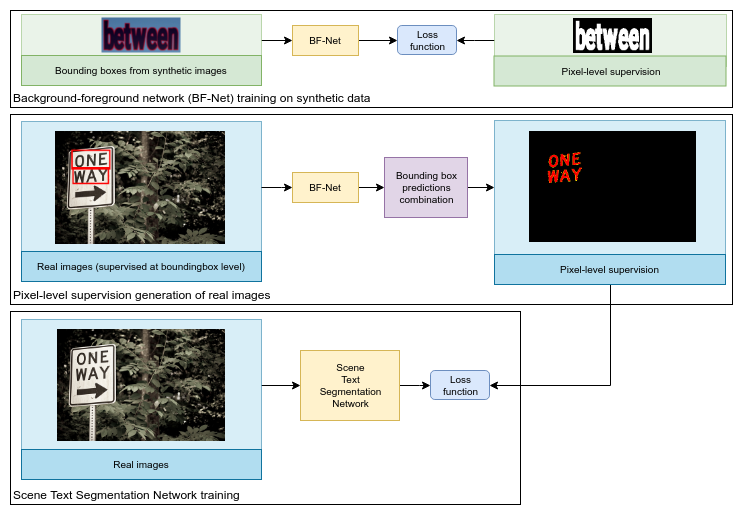}}
\caption{Scheme of the supervision generation procedure. Steps 1. to 3. are sketched starting from the top.}
\label{training_scheme}
\end{center}
\end{figure*}

Scene text segmentation is an important and challenging step in the extraction of textual information in natural images. It aims at making dense predictions in order to detect, for each pixel of an image, the presence of text. Convolutional Neural Networks (CNNs) have achieved the state of the art in many computer vision tasks, including scene text segmentation. Nonetheless, their training is usually based on large sets of fully supervised data. To the best of our knowledge, only two public datasets are available for scene text segmentation, i.e. ICDAR--2013 \cite{icdar2013} and Total--Text \cite{totalText}, that contain a number of pixel--level annotated images barely sufficient to train a deep segmentation network. 
A solution to this problem has been proposed in \cite{tang2017scene}, where a pixel--level supervision is produced employing the synthetic image generator introduced by \cite{synthText}. 
However, unfortunately, there is no guarantee that a network trained on synthetic data will generalize to real images. This usually depends on the quality of the generated data (i.e. how much they are similar to real images), since the domain--shift may affect the generalization capability of the model. \\
\noindent
In this paper, we propose to employ weak supervisions to improve the performances on real data. Indeed, a lot of datasets for text localization, in which the supervision is given by bounding--boxes around the text, are available (f.i. COCO text \cite{cocotext}, ICDAR--2013 \cite{icdar2013}, ICDAR--2015 \cite{icdar2015}, and MLT \cite{MLT}). In fact, obtaining this type of annotations is easier than providing a full pixel--level supervision, despite being less accurate. Inspired by \cite{bbox}, we adopt a training procedure (see Figure \ref{training_scheme}) that exploits weak annotations for pixel by pixel labeling. Specifically, the generation procedure consists of two distinct steps.
 \begin{enumerate} 
\item A background--foreground network is trained on a large dataset of synthetically generated images with full pixel--level supervision. The purpose of this network is to recognize text within a bounding--box. 
\item A scene text segmentation network is trained on a text localization dataset, in which the pixel--level supervision is obtained exploiting the output of the background--foreground network.
 \end{enumerate}
\noindent The logic behind this approach is that training a segmentation network focused on a bounding--box is a simpler task than using the entire image. In fact, inside a bounding--box, the text dimension is known (directly related to the box dimensions) and the background (i.e. non textual objects) variability is reduced. Moreover, the box annotation gives a precise information on the text position, since each pixel which is not included in a box does not represent text. Therefore, we exploit weak annotations to produce accurate pixel--level supervisions for a dataset of real images, which allows to reduce the domain--shift between synthetic and real data. 
In particular, employing the background--foreground network, an accurate pixel--level supervision for two datasets of real images, COCO--Text \cite{cocotext} and MLT \cite{MLT}, have been generated.
The datasets with the obtained supervision, COCO--Text--Segmentation (COCO\_TS) and MLT--Segmentation (MLT\_S), will be made publicly available\footnote{\url{http://clem.diism.unisi.it/~coco_ts/}} to foster reproducibility and to promote future research in scene text segmentation. \\
\noindent
In particular, to deal with the specificity of scene text segmentation, we introduce the Segmentation Multiscale Attention Network (SMANet), a deep fully convolutional neural network with a ResNet backbone encoder. In the SMANet architecture, a convolutional decoder is employed to recover fine details, which are lost due to the presence of pooling and strided convolutions. A multiscale attention mechanism is also used to focus on the most informative part of the image (i.e. on areas containing text). 
The proposed architecture has been employed both to implement the background--foreground network, needed for the extraction of the segmentation datasets, and for the scene text segmentation network. 
Differently, from other state--of--the--art approaches, which use complicated multistage architectures, the SMANet is trained end--to--end employing the COCO\_TS and MLT\_S datasets. An ablation study has been devised to evaluate the role of the double convolutional decoder and of the multiscale attention module in the SMANet.\\
\noindent
Furthermore, a series of experiments were conducted to evaluate the effectiveness of the proposed datasets compared to the use of synthetic data, as previously proposed in literature.
The obtained results suggest that, using both the COCO\_TS and the MLT\_S datasets, a deep convolutional segmentation network can be trained more efficiently than using synthetic data, employing only a fraction of the learning set. \\
\noindent
The paper is organized as follows. In Section \ref{Related_Works}, related works are briefly reviewed. In Section \ref{method}, the weakly segmentation generation procedure, used for COCO\_TS and MLT\_S, is described and the SMANet architecture, tailored to scene text segmentation, is presented. Section \ref{Experiments} reports the experimental setup and the results obtained on the ICDAR--2013 and Total--Text datasets. Finally, some conclusions are drawn in Section \ref{Conclusions}.

\section{Related Works} \label{Related_Works}
The proposed method is related to five main research topics, namely synthetic data generation, bounding--boxes for semantic segmentation, semantic segmentation with CNNs, scene text segmentation, and attention modules, whose literature is reviewed in the following.
\paragraph{Synthetic data generation} 
Synthetic datasets are a cheap and scalable alternative to the human ground--truth supervision in machine learning. Recently, several papers reported on the use of synthetic data to face a variety of different problems. Large collections of synthetic images of driving scenes in urban environments were generated in \cite{SYNTHIA}, synthetic indoor scenes have been exploited by \cite{indoor}, while artificial images of Petri plates were created in \cite{synth_colony}. In text analysis, the use of synthetic data for text spotting, localization and recognition has been proposed in \cite{jaderberg_text}. Moreover, an improved synthetic data generator for text localization in natural images was introduced by \cite{synthText}. The engine is designed to overlay text strings in existing background images. The text is rendered in image regions characterized by an uniform color and texture, also taking into account the 3D geometry of the scene.
 This synthetic data generator engine has been modified in \cite{tang2017scene} to extract pixel--level annotations. Similarly to \cite{tang2017scene}, in this work, the engine proposed in \cite{synthText} was used for scene text segmentation.

\paragraph{Bounding--boxes for semantic segmentation} 
In order to reduce the data labeling efforts, weakly supervised approaches aim at learning from weak annotations, such as image--level tags, partial labels, bounding--boxes, etc. The bounding--box supervision was used to aid semantic segmentation in \cite{BoxSup}, where the core idea is that to iterate between automatically generating region proposals and training convolutional networks. Similarly, in \cite{EM_weak}, an Expectation--Maximization algorithm was used to iteratively update the training supervision. Instead, in \cite{bbox_khoreva}, a GrabCut--like algorithm is employed to generate training labels from bounding boxes. Finally, more related to this work, in \cite{bbox}, the segmentation supervision for a semantic segmentation network is directly produced from bounding--box annotations, exploiting a deep CNN.

\paragraph{Semantic segmentation with CNNs} 
Image semantic segmentation aims at inferring the class of each pixel of an image. Recent semantic segmentation algorithms often convert existing CNN architectures, designed for image classification, to fully convolutional networks \cite{FCN}. Normally, these networks have an encoder--decoder structure. Moreover, the level of details required by semantic segmentation inspired the use of dilated convolution to enlarge the receptive field without decreasing the resolution \cite{atrous}.  
Besides, different solutions have been proposed to deal with the presence of objects at different scales. The Pyramid Scene Parsing Network (PSPNet) \cite{PSP} applies a pyramid of pooling to collect contextual information at different scales. Instead, DeepLab \cite{deeplab} employs atrous spatial pyramid pooling, which consists of parallel dilated convolutions with different rates.

\paragraph{Scene Text Segmentation} 
Document image segmentation has a long history and was originally based on thresholding approaches (local \cite{laplacian}, global \cite{binarization} or adaptive \cite{otsu}). The application of these methods to scene text segmentation is quite challenging, due to the high variability of conditions that can be found in natural images. To face this variability, in \cite{bai2014seed}, low level features are used to identify the seed points of texts and backgrounds and then to segment the text using semi--supervised learning. In \cite{mishra2011mrf}, the binarization of scene text has been formulated as a Markov Random Field model optimization problem, where the optimal binarization is obtained iteratively with Graph Cuts. To improve the segmentation performance, a multilevel Maximally Stable Extremal Region (MSER) approach was presented in \cite{tian2014scene}. The MSER strategy is applied together with a text candidate selection algorithm based on hand--extracted text--specific features. Finally, in \cite{tang2017scene}, a CNN approach to scene text segmentation is described, which employs three stages for extraction, refinement and classification. Based on the seminal paper \cite{coco_ts}, the present work extends the pixel--level annotation generation procedure to the MLT dataset, also introducing a new convolutional neural network architecture (SMANet) specific to scene text segmentation.
\paragraph{Attention modules}
The capacity of learning long--range dependencies have popularized the use of attention modules in a wide variety of Natural Language Processing (NLP) tasks. In particular, the use of a self--attention mechanism to extract global dependencies in machine translation was originally proposed in \cite{vaswani}. After that, many other NLP applications have employed attention mechanism \cite{shen,lin2017}.
Attention modules have been also increasingly applied in computer vision.
In \cite{SE_Networks}, an attention mechanism is proposed to model the interdependencies between channels. In \cite{zhang}, self--attention modules are employed to improve image generation. An attention mechanism is used in \cite{chen2016attention} to weigh multiscale features extracted by a shared network.
Contextual information is captured by a self--attention mechanism in OCNet \cite{ocnet} and DANet \cite{danet}.
In PSA \cite{psanet}, an attention map is learned to aggregate contextual information adaptively for each individual point.
Inspired by the success of attention mechanisms in computer vision, we introduce an attention module to gather context information and focus on text.

\section{Materials and Methods}\label{method}
In the following, a general overview of the proposed method is provided. The sets of data involved in the creation of the COCO\_TS and MLT\_S datasets are introduced in Section \ref{Benchmark Datasets}. Section \ref{COCO_Text_and_MLT_Segmentation} describes the weakly supervised approach used to generate COCO\_TS and MLT\_S, while, in Section \ref{Scene_Text_Segmentation}, the generated supervision of the two datasets is used to train a deep segmentation network. Finally, Section \ref{smanet} presents the architectural details of the Segmentation Multiscale Attention Network (SMANet).

\subsection{Datasets} \label{Benchmark Datasets}

\paragraph{Synthetic dataset} 
In this work, the same generation process proposed by \cite{synthText} has been employed to create a large set of synthetic scene text images. The engine renders synthetic text to existing background images, accounting for the local three dimensional scene geometry. A synthetic dataset of about 800,000 images was generated following this procedure. 
From this set of images, about 1,000,000 image crops have been extracted. Specifically, for each word, a bounding--box is defined and enlarged by a factor of 0.3, and then the image is cropped around the bounding--box. These bounding boxes have been used to train the background--foreground network described in Section \ref{COCO_Text_and_MLT_Segmentation}.

\paragraph {COCO--Text}
The COCO--2014 dataset \cite{coco}, firstly released by Microsoft Corporation, collects instance--level fully annotated images of natural scenes. COCO--Text \cite{cocotext} is based on COCO--2014 and contains a total of 63,686 images, split in 43,686 training, 10,000 validation and 10,000 test images, supervised at the bounding--box level for text localization. Differently from other scene text datasets, the COCO--2014 dataset was not collected specifically for the extraction of textual information, hence some of its images do not contain text. Therefore, for the generation of the proposed COCO\_TS dataset, a subset of 14,690 images have been selected from COCO--Text, each one including at least a bounding--box labeled as legible, machine printed, and written in English.

\paragraph{MLT}
The MLT dataset \cite{MLT} has been collected for the ICDAR--2017 \cite{icdar2017} competition and comprises natural scene images with embedded text, such as street signs, street advertisement boards, shops names, passing vehicles and users photos in microblogs. The images have been captured by different users', using various mobile phone cameras. A large percentage of the images contains more than one language. 
The dataset is composed by 18,000 images containing text of nine different languages all annotated at the bounding--box level.

\paragraph {ICDAR--2013}
The ICDAR--2013 \cite{icdar2013} dataset collects a training and a test set containing 229 and 233 images, respectively. The images are extracted from ICDAR--2011 \cite{icdar2011}, after the removal of duplicated images and with some revisited ground--truth annotations. The scene text segmentation challenge in the ICDAR--2015 competition \cite{icdar2015} was based on the same datasets as ICDAR--2013.

\paragraph {Total--Text}
Total--Text \cite{totalText} is a scene text dataset which collects 1255 training images and 300 test images with a pixel--level supervision. Differently from ICDAR--2013, where texts have always a horizontal appearance, this dataset contains images with texts showing highly diversified orientations.

\subsection{COCO\_TS and MLT\_S Datasets}\label{COCO_Text_and_MLT_Segmentation}
\begin{figure}[!ht]
\begin{center}
\includegraphics[width=3.5cm,keepaspectratio]{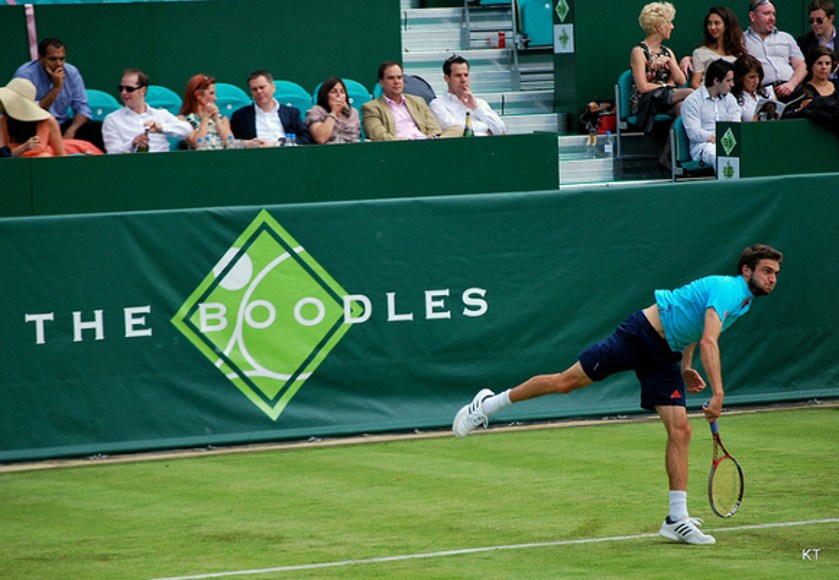}
\includegraphics[width=3.5cm,keepaspectratio]{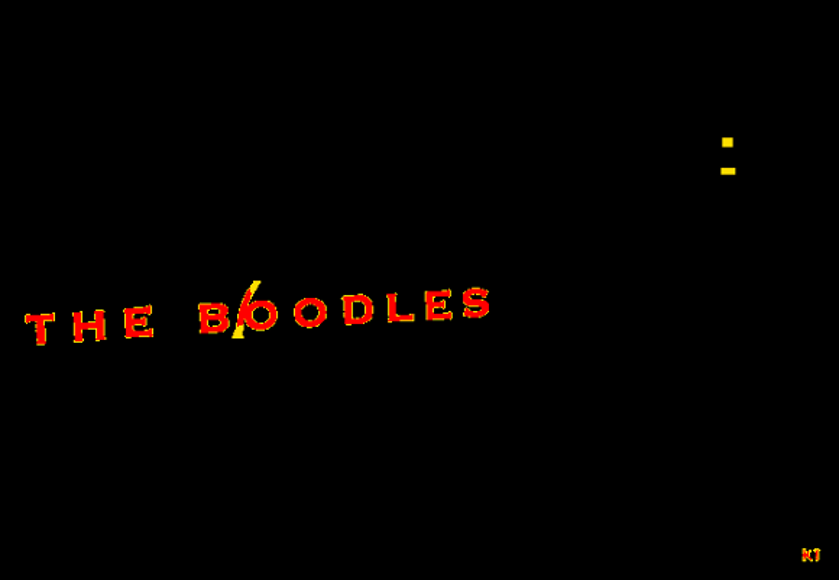}
\includegraphics[width=3.5cm,keepaspectratio]{generated_supervision/COCO/COCO_train2014_000000025102.jpg}
\includegraphics[width=3.5cm,keepaspectratio]{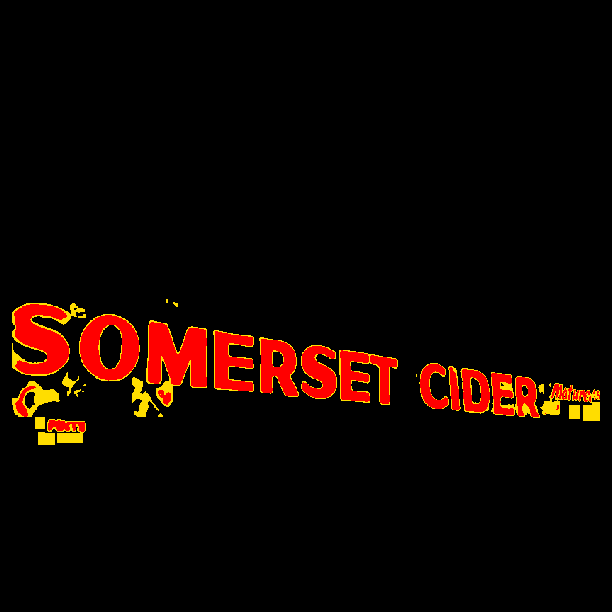}
\includegraphics[width=3.5cm,keepaspectratio]{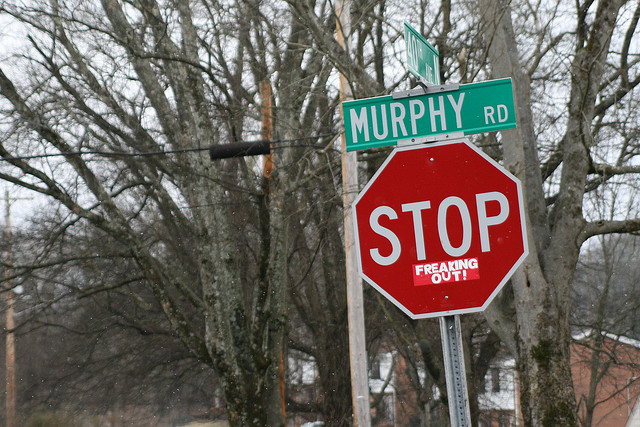}
\includegraphics[width=3.5cm,keepaspectratio]{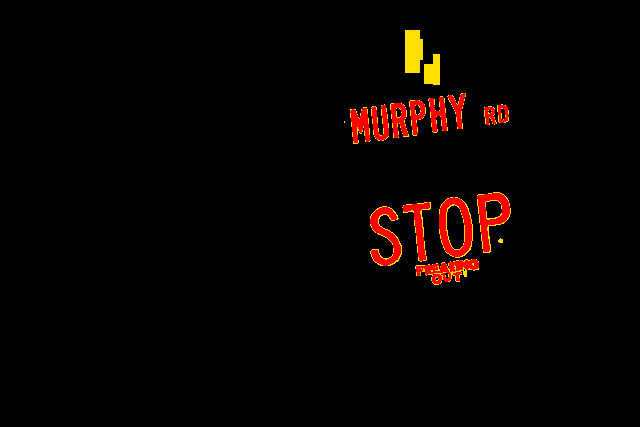}
\caption{The original images and the generated supervisions of the COCO\_TS dataset, on the top and at the bottom, respectively. The background is colored in black, the foreground in red, and the uncertainty region in yellow.}
\label{supervision_example_coco_ts}
\end{center}

\begin{center}
\includegraphics[width=3.5cm,keepaspectratio]{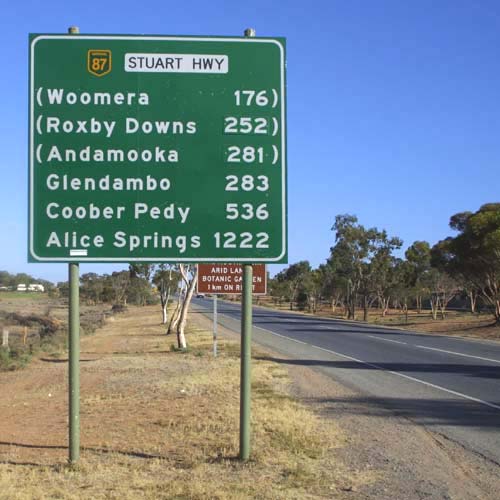}
\includegraphics[width=3.5cm,keepaspectratio]{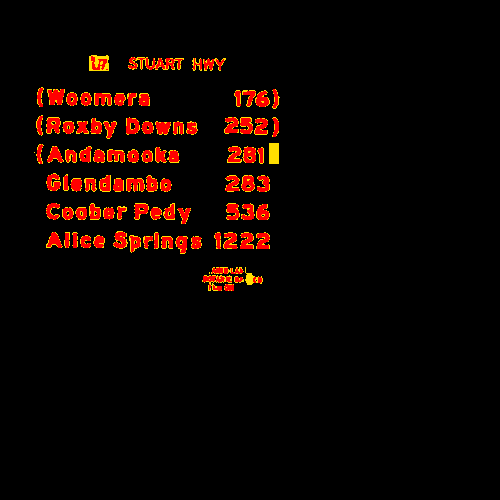}
\includegraphics[width=3.5cm,keepaspectratio]{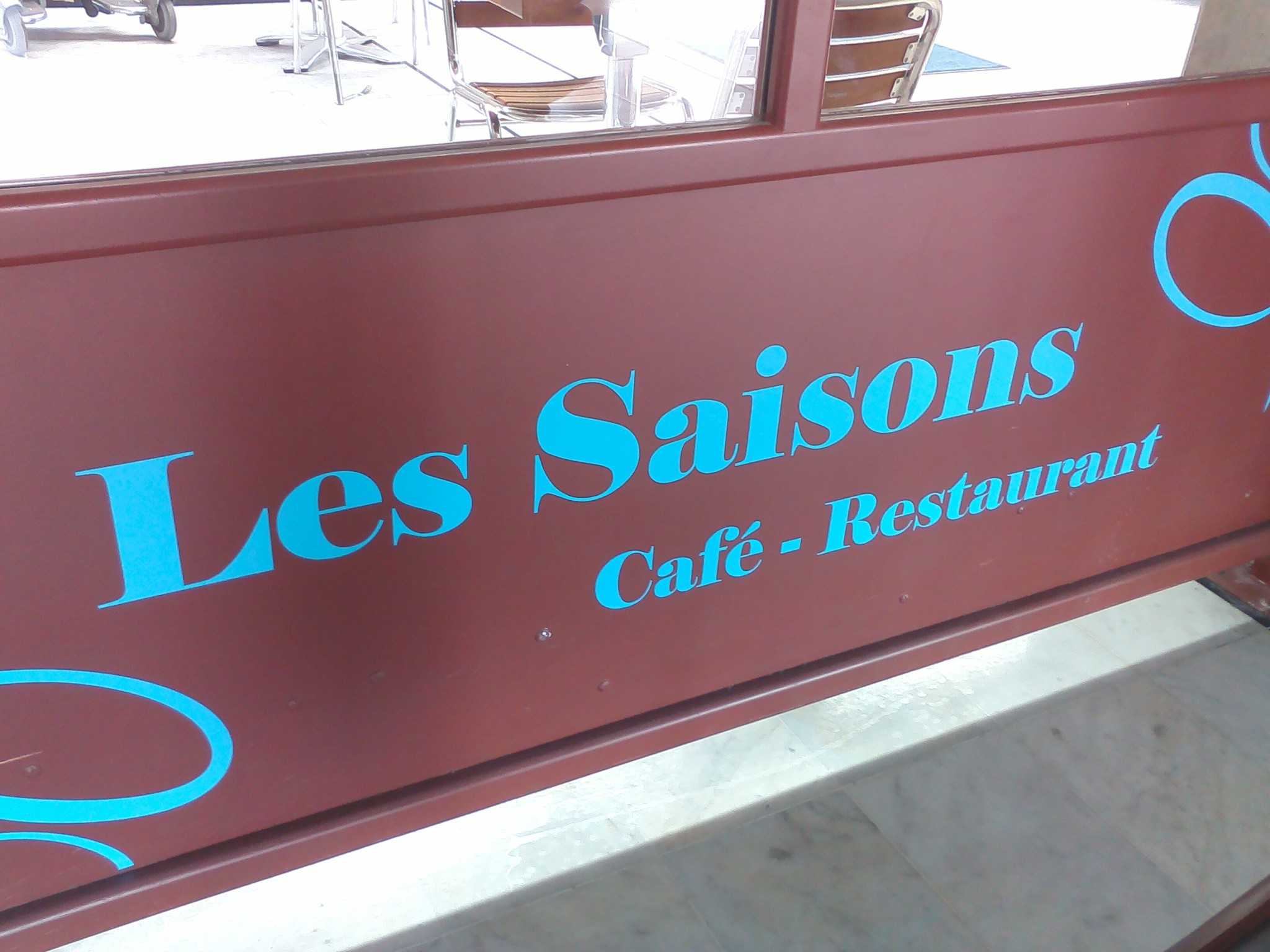}
\includegraphics[width=3.5cm,keepaspectratio]{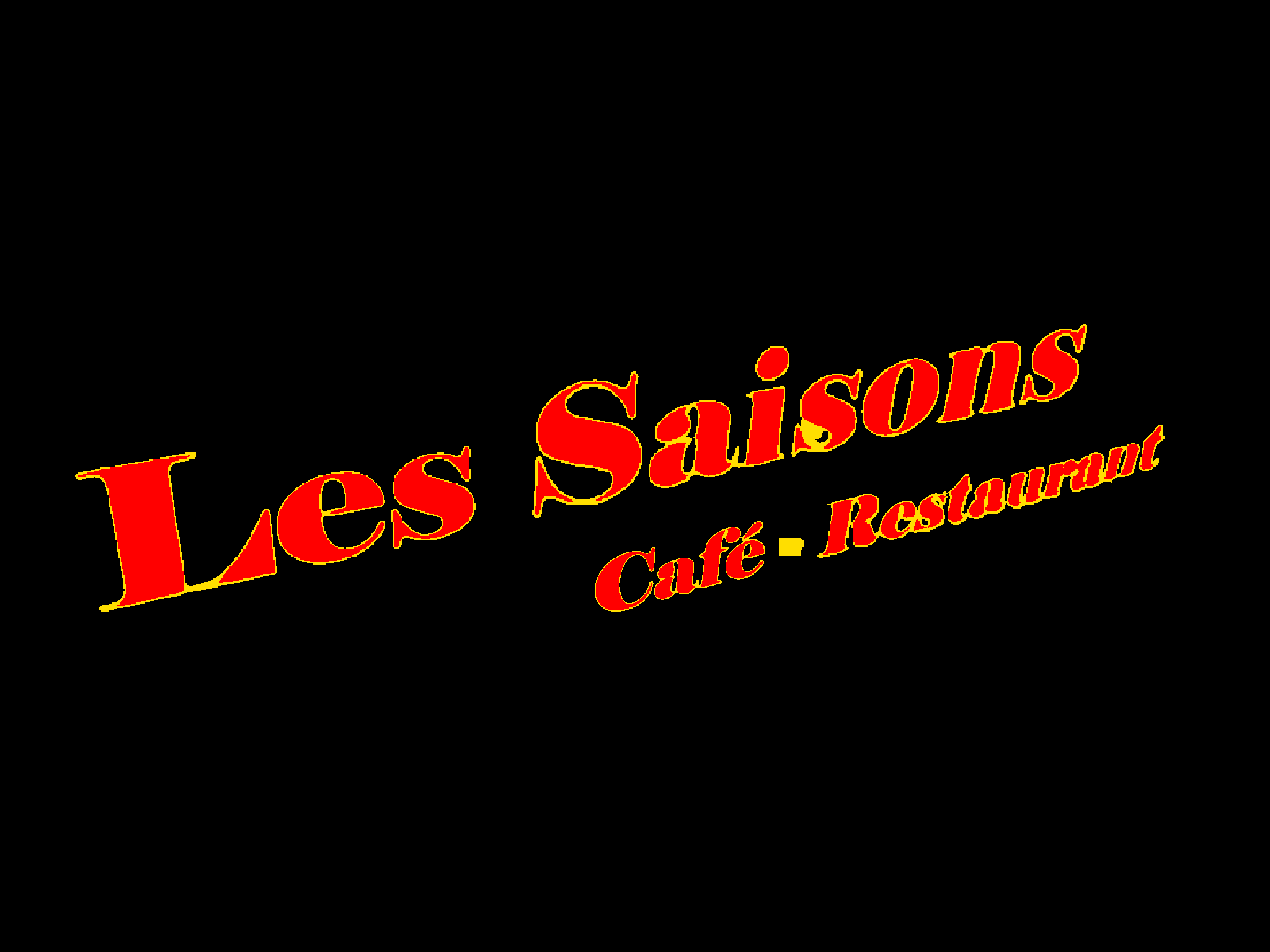}
\includegraphics[width=3.5cm,keepaspectratio]{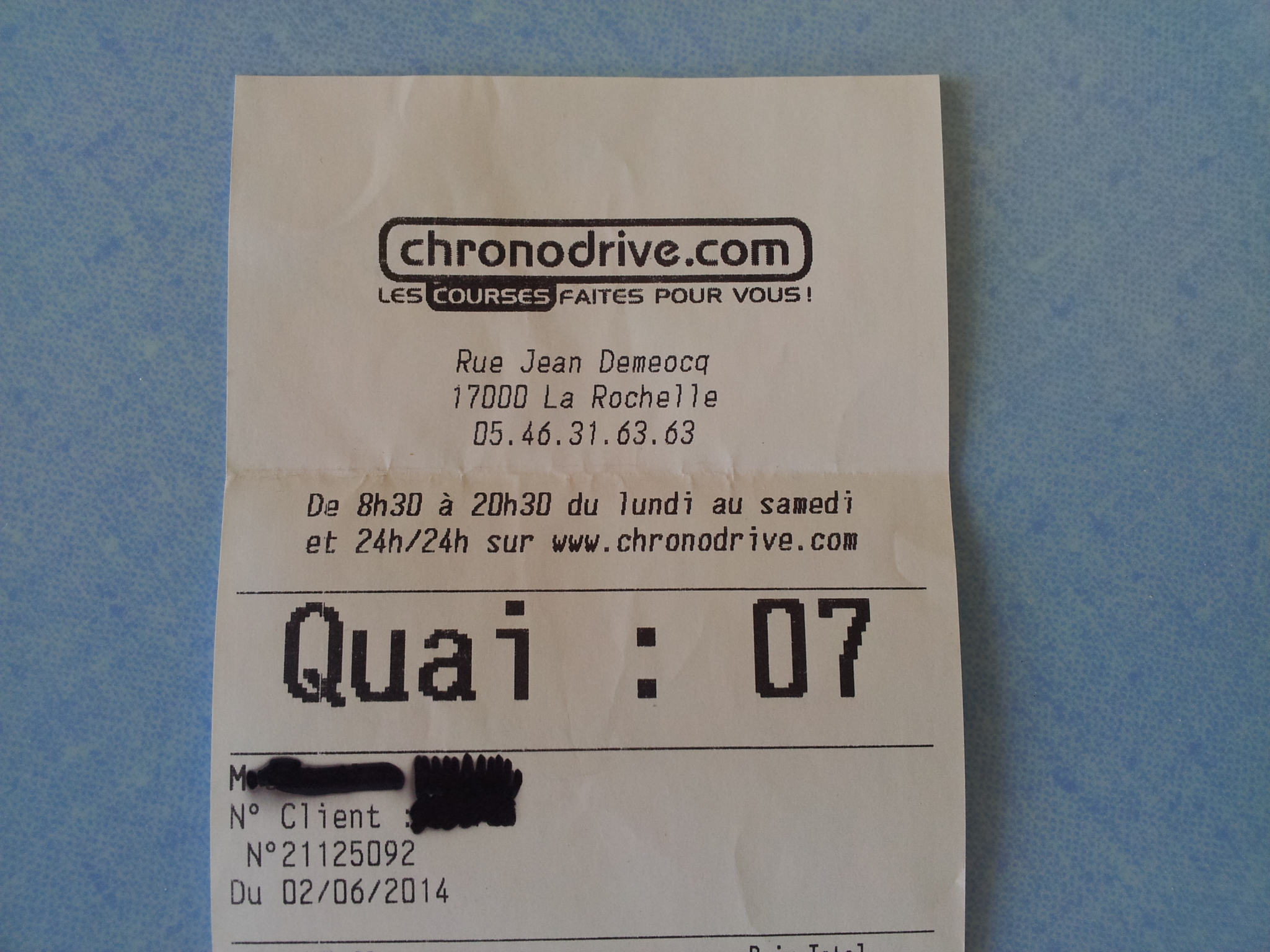}
\includegraphics[width=3.5cm,keepaspectratio]{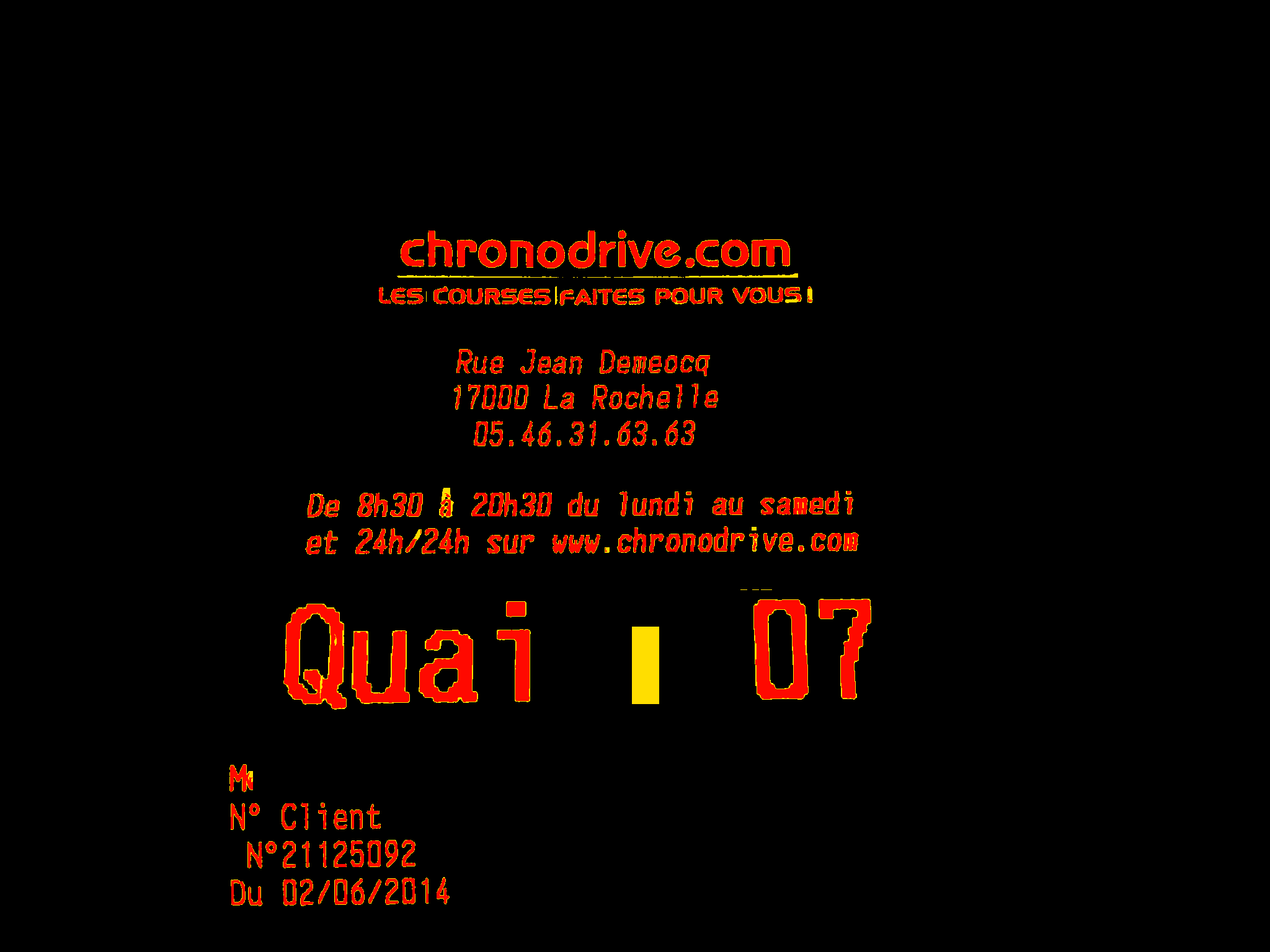}
\caption{The original images and the generated supervisions of the MLT\_S dataset on the top and at the bottom, respectively. The background is colored in black, the foreground in red, and the uncertainty region in yellow.}
\label{supervision_example_mlt_s}
\end{center}
\end{figure}

Collecting supervised images for scene text segmentation is costly and time consuming. In fact, only few datasets with a reduced number of images are available. Instead, numerous datasets provide bounding--box level annotations for text detection. In this paper, we introduce the COCO\_TS dataset, which provides 14,690 pixel--level supervisions for the COCO--Text images, and the MLT\_S dataset, which contains 6896 (5540 from the training set and 1356 from the validation set) label maps for the MLT images. The supervision is obtained from the bounding--boxes, available in both the original datasets, exploiting a weakly supervised algorithm. The supervision generation procedure, summarized in Figure \ref{training_scheme}, consists of three different steps.
\noindent
\begin{enumerate} 
\item A background--foreground network is trained on synthetic data to extract text from bounding--boxes. 
\item The background--foreground network is employed to generate pixel--level supervisions for real images of both the COCO--Text and the MLT datasets. 
\item A scene text segmentation network is trained on the real images with the generated supervisions.
 \end{enumerate} 
In particular, a deep neural network was trained to segment the text inside a bounding--box, thus separating the background from the foreground. The rationale beneath the proposed approach is that realizing a background--foreground segmentation, constrained to a bounding--box, is significantly simpler than producing the segmentation of the whole image. For this reason, we suppose that, even if trained on synthetic data, the background--foreground network can effectively be used to segment text in bounding boxes extracted from real images. To train the background--foreground network, pixel--level supervisions of a significant number of bounding--boxes is required. The 1,000,000 bounding--box crops extracted from the synthetic dataset have been used to this purpose. 
After the training phase, the background--foreground network is applied on the bounding--boxes extracted from the COCO--Text and the MLT datasets.
For each image, the pixel--level supervision is obtained combining the probability maps (calculated by the background--foreground network) for all the bounding--boxes inside the image. It can happen that a region belongs to more than one bounding box (f.i., two written texts close to each other could have overlapping bounding--boxes) and, in this case, the prediction with the highest foreground probability value is considered. The final pixel--wise annotation $l(x, y)$, at position $(x, y)$, is obtained employing two fixed thresholds, $th_1$ and $th_2$, on the probability maps $prob(x,y)$:
  \begin{equation}
    l(x, y)=\left\{
                \begin{array}{ll}
                 background \quad \quad \textrm{if} \quad prob(x, y) < th_1 \\
                 foreground \quad \quad \textrm{if} \quad prob(x, y) > th_2\\
                  uncertain \quad \quad \quad \hspace*{-1mm}  \textrm{otherwise}
                \end{array}
              \right.
\label{FixedTh}
\end{equation}
The two thresholds $th_1$ and $th_2$ have been fixed to $0.3$ and $0.7$, respectively, based on a grid search approach. 
If $prob(x, y) \in (th_1,th_2)$, then $(x,y)$ is labeled as uncertain. To provide a significant pixel--level supervision, bounding--boxes that are not labeled as legible, machine printed and written with Latin characters have been added to the uncertainty region. The insertion, in the generated supervision of this uncertainty region, has prove to be effective, avoiding the gradient propagation in regions where text could be potentially misclassified with the background. This procedure has been used to extract the COCO\_TS and the MLT\_S datasets.
Some examples of the obtained supervisions are reported in Figure \ref{supervision_example_coco_ts} and \ref{supervision_example_mlt_s}.

\begin{figure*}[!ht]
\vskip 0.2in
\begin{center}
\centerline{\includegraphics[scale=0.4]{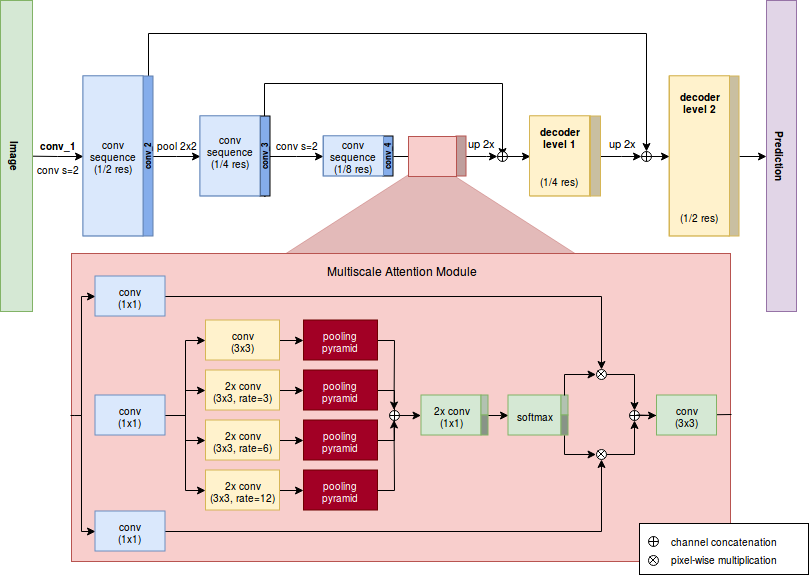}}
\caption{On the top, the overall network architecture. At the bottom, the details of the multiscale attention module.}
\label{network_architecture}
\end{center}
\end{figure*}

\subsection{Scene Text Segmentation} \label{Scene_Text_Segmentation}
The COCO\_TS and the MLT\_S datasets are used to train a deep segmentation network (bottom of Figure \ref{training_scheme}) for scene text segmentation of both the ICDAR--2013 and the Total--Text datasets. The effects obtained by the use of the generated datasets, as an alternative to synthetic data, will be described in Section \ref{Experiments}. 

\subsection{SMANet Architecture}\label{smanet}
To deal with the specificity of scene text segmentation we also propose the Segmentation Multiscale Attention Network, which is composed by three main components: a ResNet encoder, a multiscale attention module, and a convolutional decoder.
The encoder is based on the PSPNet for semantic segmentation, which is a deep fully convolutional neural network that re--purposes the ResNet \cite{ResNet}, originally designed for image classification. In the PSPNet, a set of dilated convolutions (i.e. atrous convolution \cite{atrous}) replaces standard convolutions in the ResNet backbone, to enlarge the receptive field of the neural network. To gather context information, the PSP exploits a pyramid of pooling with different kernel size.
In this work, the ResNet50 architecture is used as the CNN encoder.
Despite the PSPnet proved to be very effective in the segmentation of natural images, scene text segmentation is a specific task with its own peculiarities. In particular, text in natural images can have a high variability of scales and dimensions. To better handle the presence of thin text, similarly to \cite{tang2017scene}, we modified the network structure adding a two level convolutional decoder.
A multiscale attention mechanism is also employed to focus on the text present in the image. 
The overall SMANet architecture is depicted in Figure \ref{network_architecture}, and detailed in the following.

\subsubsection{Multiscale Attention Module}

\begin{figure}[!ht]
\vskip 0.2in
\begin{center}
\includegraphics[width=3.5cm,keepaspectratio]{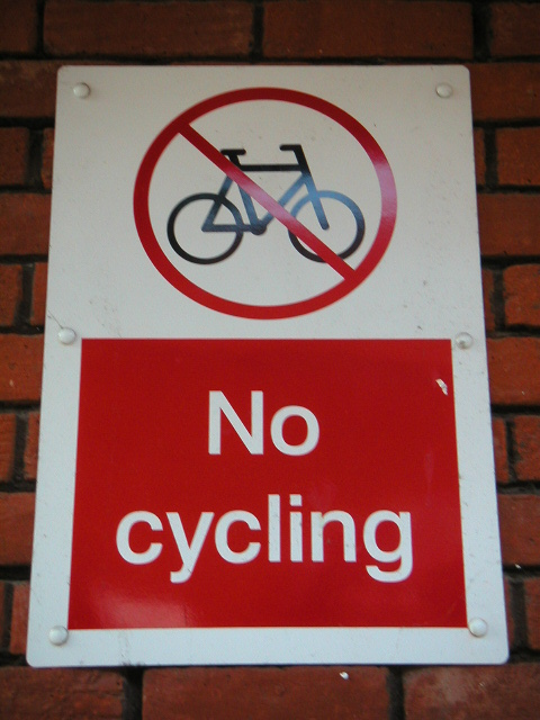}
\includegraphics[width=3.5cm,keepaspectratio]{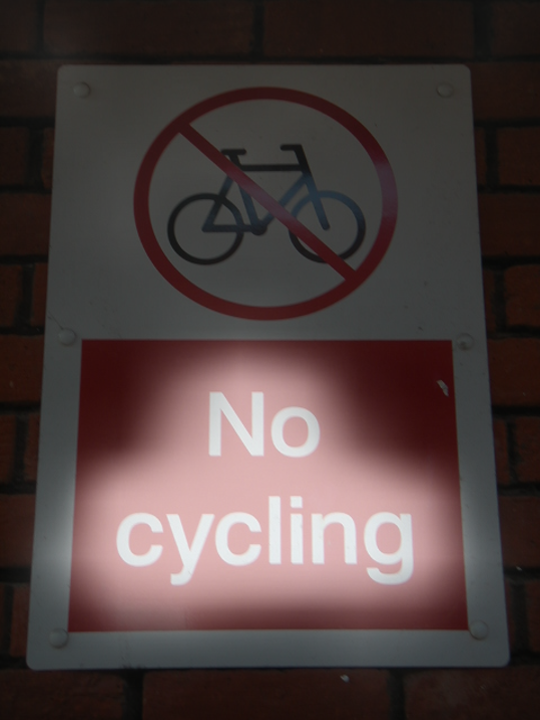}
\includegraphics[width=3.5cm,keepaspectratio]{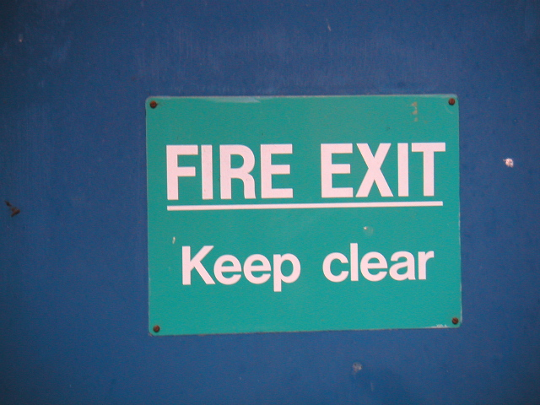}
\includegraphics[width=3.5cm,keepaspectratio]{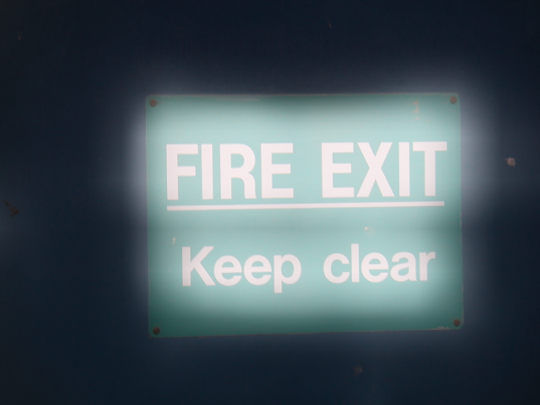}
\includegraphics[width=3.5cm,keepaspectratio]{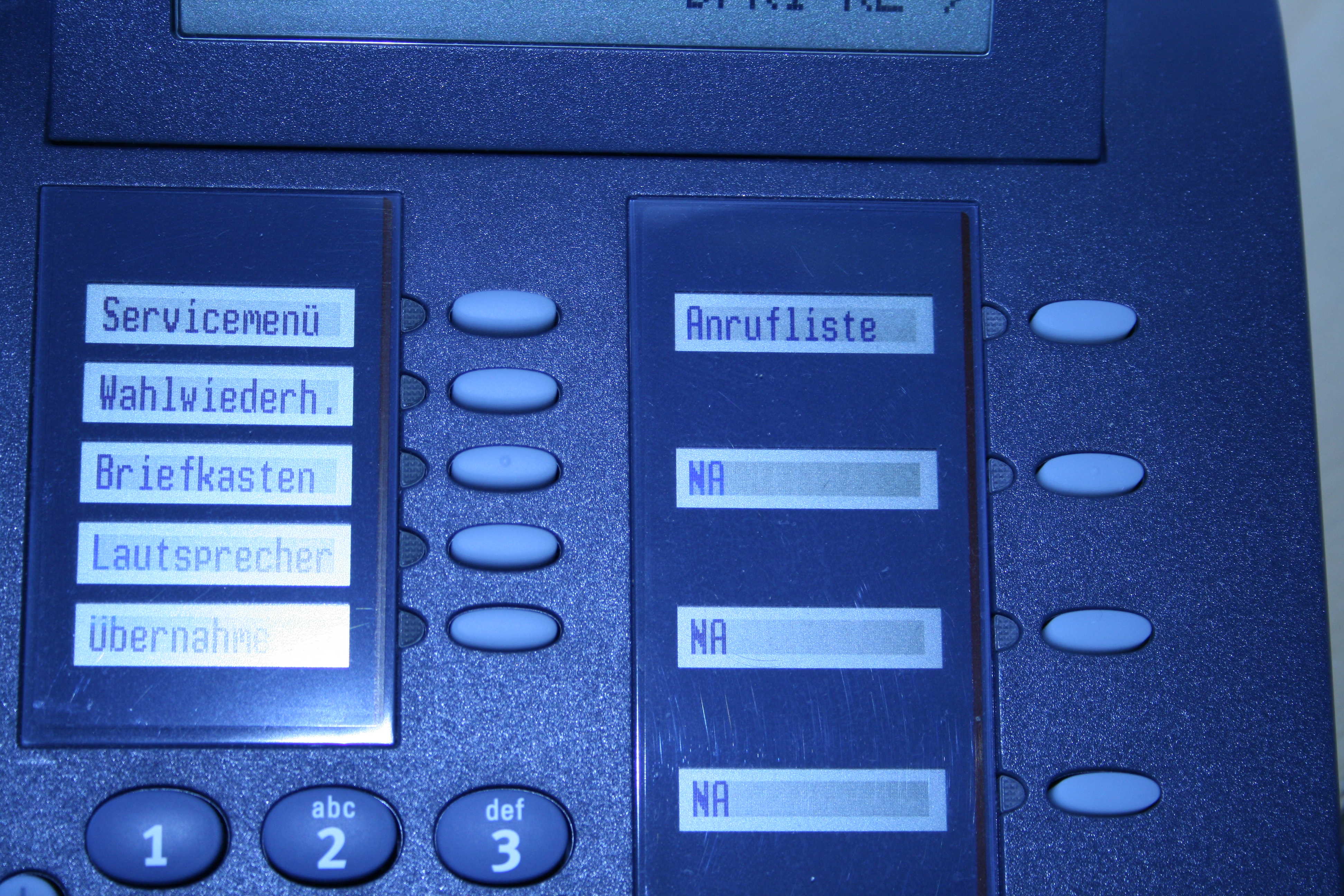}
\includegraphics[width=3.5cm,keepaspectratio]{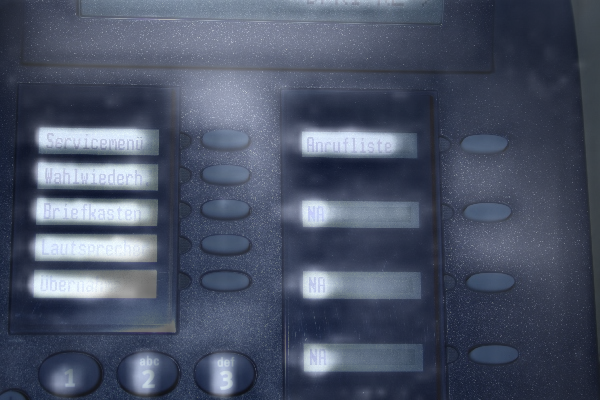}
\includegraphics[width=3.5cm,keepaspectratio]{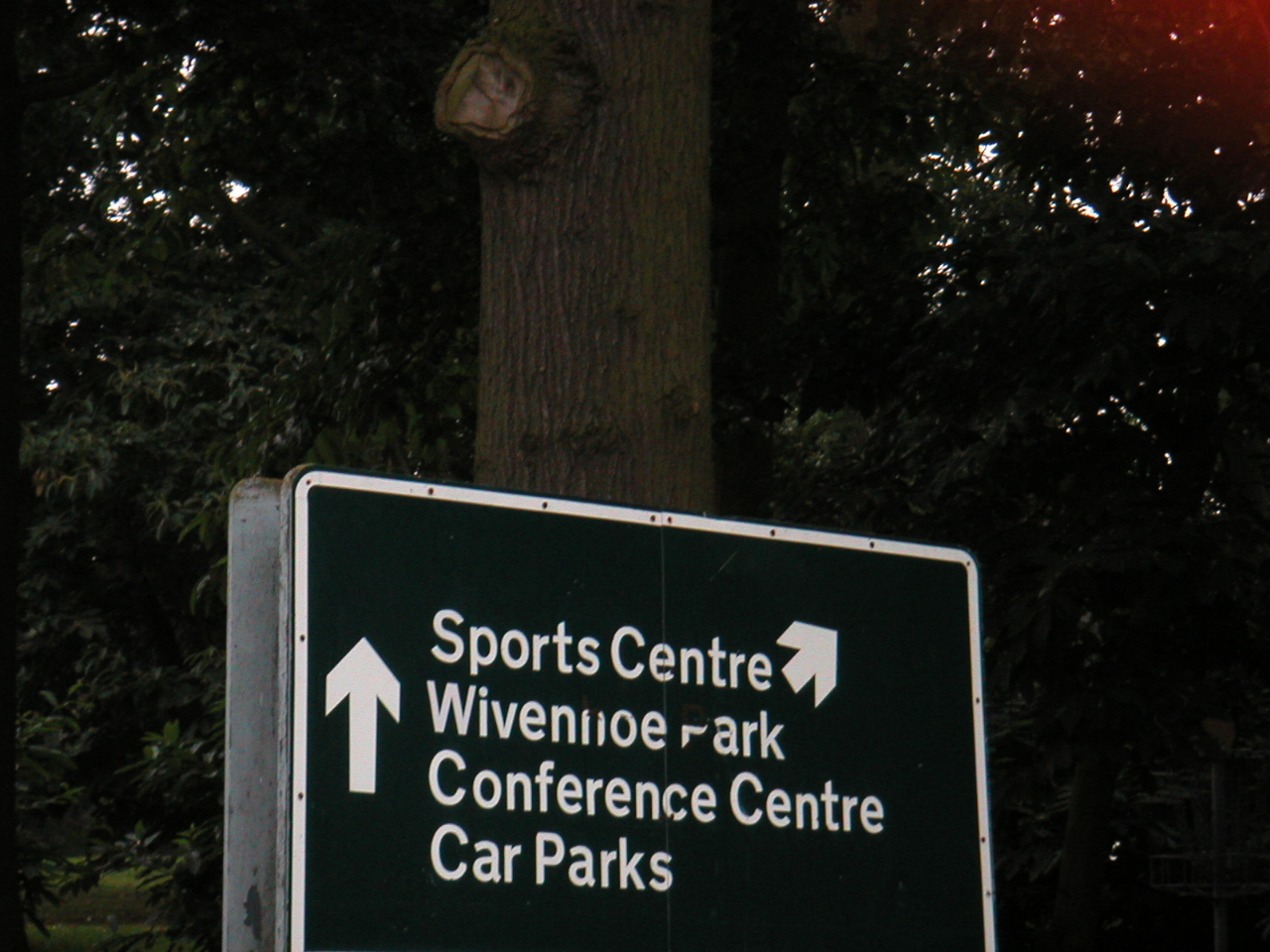}
\includegraphics[width=3.5cm,keepaspectratio]{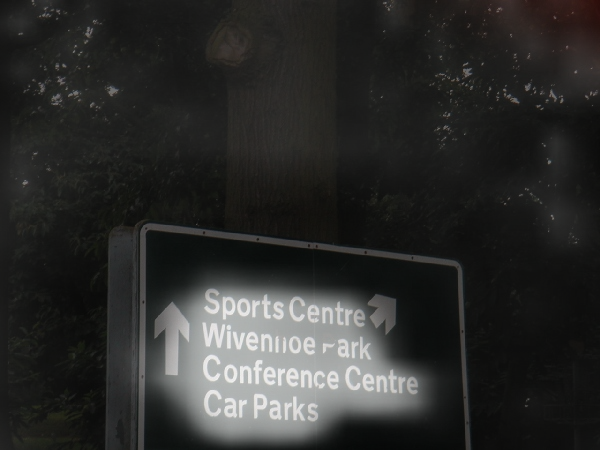}
\caption{The original images (first row) and their attention maps (second row).}
\label{attention_examples}
\end{center}
\end{figure}

Recent state--of--the--art models, such as the PSPNet \cite{PSP} or DeepLab \cite{deeplab,deeplab_v3}, employ a Spatial Pyramid Pooling (PSP) or an atrous spatial pyramid pooling module to gather information at different scales.
Dilated convolutions may be harmful for the local consistency of feature maps, whereas the PSP module loses pixel precision during the different scales of pooling operations.
To address this problem, we introduce a pooling attention mechanism that provides pixel--level attention for the features extracted by the ResNet encoder.
The architectural details of the attention module are depicted in Figure \ref{network_architecture}. 
Specifically, the CNN encoded representation passes through a convolutional layer and then it is given as input to the SMANet attention module that outputs the attention maps. These maps are pixel--wise multiplied with the CNN encoded representation, previously passed through a 1 by 1 convolutional layer to conveniently reduce the feature map dimensions. 
The SMANet attention module is composed by a pyramid of atrous convolutions at different dilation rate, each one followed by a PSP module. The obtained multiscale representation is concatenated and given as input to a couple of convolutional layers and a softmax, that output two attention maps.
The SMANet attention module is followed by a two level decoder to recover small details at a higher resolution.
Figure \ref{attention_examples} shows some examples of the attention maps produced by the proposed multiscale attention module. In particular, it can be observed that the attention mechanism learns to focus on regions containing text.

\subsubsection{Decoder}

The low dimensional feature maps, "conv\_2" and "conv\_3" (see Figure \ref{network_architecture}), are the last convolutional layers of the ResNet encoder at, respectively, 1/2 and 1/4 of the original resolution.
A $1\times1$ convolution is applied on these low level feature maps to reduce the number of channels. The SMANet attention module output is first bilinearly upsampled by a factor of 2 and then concatenated with the corresponding low level features at the same spatial resolution (conv\_2). Then, the concatenated feature maps are followed by two convolutional layer and upsampled again by a factor of 2. The resulting feature maps are concatenated with conv\_3 and, finally, two convolutional layers and a last upsample produce the pixel--wise prediction.

\section{Experiments} \label{Experiments}
In the following, our experimental setup is described. In particular, Section \ref{training_details} introduces the implementation details used in our experiments, whereas Section \ref{sts_eval} discusses the results achieved using the COCO\_TS and the MLT\_S datasets.
Finally, in Section \ref{smanet_eval} an ablation study is presented to show the advantages introduced by the use of the SMANet architecture in scene text segmentation.

\subsection{Implementation Details}\label{training_details}
The SMANet is implemented in TensorFlow. 
The CNN encoder of the two networks exploits the ResNet50 model. 
The experiments were realized based on the training procedure explained in the following.
As far as the background--foreground network is considered, the image crops were resized to have a min side dimension of 185, while maintaining the original aspect--ratio. Random crops of $185\times185$ were used during training. Instead, for the scene text segmentation network, the input images were not resized, and random crops of $281\times281$ were extracted for training. A multi--scale approach is employed during training and test. In the evaluation phase, a sliding window strategy was used for both the networks. The Adam optimizer \cite{adam}, with a learning rate of $10^{-4}$, was used to train the network. The experimentation was carried out in a Debian environment, with a single NVIDIA GeForce GTX 1080 Ti GPU.

\subsection{Scene Text Segmentation evaluation}
\label{sts_eval}
Due to the inherent difficulties in collecting large sets of pixel--level supervised images, only few public datasets are available for scene text segmentation. To face this problem, in \cite{tang2017scene}, synthetic data generation has been employed. Nevertheless, due to the domain--shift, there is no guarantee that a network trained on synthetic data would generalize well also to real images. 
The COCO\_TS and the MLT\_S datasets actually contain real images and, therefore, we expect that, when used for network training, the domain--shift can be reduced. To test this hypothesis, the SMANet was used for scene text segmentation and the ICDAR--2013 and Total--Text test sets, that provides pixel--level annotations, were used to evaluate the performances. In particular, we compared the following experimental setups:
 \begin{itemize} 
\item \textbf{Synth:} The training relies only on the synthetically generated images;
\item \textbf{Synth + COCO\_TS:} The network is pre--trained on the synthetic dataset and fine--tuned on the COCO\_TS images;
\item \textbf{Synth + MLT\_S:} The synthetic dataset is used to pre--train the network and the MLT\_S images are used for fine--tuning;
\item \textbf{COCO\_TS:} The network is trained only on the COCO\_TS dataset.
\item \textbf{MLT\_S:} The network is trained only on the MLT\_S dataset.
\item \textbf{COCO\_TS + MLT\_S:} The network is trained with both the COCO\_TS and the MLT\_S datasets.  
 \end{itemize} 
\noindent The influence of fine--tuning on the ICDAR--2013 and Total--Text datasets was also evaluated. The results, measured using the pixel--level precision, recall and F1 score, are reported in Table \ref{ICDAR2013_Results} and Table \ref{TotalText_Results}, respectively.

\begin{table*}[!ht]
\label{icdar_results}
\begin{center}
\begin{small}
\subfloat[\small{Results on the ICDAR--2013 test set} \label{ICDAR2013_Results}]{%
\begin{tabular}{lcccc}
\hline
 & Precision & Recall & F1 Score & \\
\hline
Synth & 73.30\%& 57.30\% & 64.28\% & --\\
Synth + COCO\_TS & 80.80\% & 71.2\% & 75.69\% & +11.41\% \\
Synth + MLT\_S & 81.70\%& 81.8\% & 81.77\% & +17.49\% \\
COCO\_TS & 80.40\% & 73.00\% & 76.52\% & +12.24\%\\
MLT\_S & 82.70\% & 80.50\% & 81.59\% & +17.31\% \\
\textbf{COCO\_TS + MLT\_S} & 82.80\% & 81.40\% & \textbf{82.10\%} & \textbf{+17.82\%} \\
\hline
ICDAR--2013 & 72.60\%& 70.10\% & 71.31\% & --\\
\hline
Synth + ICDAR--2013 & 81.12\%& 76.00\% & 78.48\% & --\\
Synth + COCO\_TS + ICDAR--2013  & 84.00\%& 77.80\% & 80.80\% & +1.52\%\\
Synth + MLT\_S + ICDAR--2013  & 88.20\%& 80.00\% & 83.89\% & +5.41\%\\
COCO\_TS + ICDAR--2013  & 84.40\% & 78.70\% & 81.47\% & +2.99\%\\
MLT\_TS + ICDAR--2013  & 88.70\% & 80.10\% & 84.14\% & +5.66\%\\
\textbf{COCO\_TS + MLT\_TS + ICDAR--2013}  & 87.30\% & 84.40\% & \textbf{85.82\%} & \textbf{+7.34\%}\\
\hline
\end{tabular}}
\end{small}

\begin{small}
\subfloat[\small{Results on the Total--Text test set} \label{TotalText_Results}]{%
\begin{tabular}{lcccc}
\hline
 & Precision & Recall & F1 Score & \\
\hline
Synth & 56.80\%& 31.70\% & 40.67\% & --\\
Synth + COCO\_TS & 74.20\%& 59,90\% & 66.26\% & +25.59\% \\
Synth + MLT\_S & 73.10\%& 61.60\% & 66.87\% & +26.20\% \\
COCO\_TS & 74.00\% & 59.30\% & 65.80\% & +25.13\%\\
MLT\_S & 74.70\%& 60.00\% & 66.56\% & +25.89\% \\
\textbf{COCO\_TS + MLT\_S} & 73.90\% & 64.70\% & \textbf{69.00\%} & \textbf{+28.33\%} \\
\hline
Total Text & 83.40\%& 66.70\% & 74.14\% & --\\
\hline
Synth + Total Text & 84.80\%& 70.50\% & 76.97\% & --\\
Synth + COCO\_TS + Total Text  & 83.90\%& 70.50\% & 76.58\% & -0.39\%\\
Synth + MLT\_S + Total Text  & 85.50\%& 71.50\% & 77.87\% & +0.90\%\\
COCO\_TS + Total Text  & 84.80\% & 69.70\% & 76.46\% & -0.51\%\\
MLT\_S + Total Text  & 86.60\%& 70.40\% & 77.63\% & +0.66\%\\
\textbf{COCO\_TS + MLT\_S + Total Text}  & 86.00\% & 71.50\% & \textbf{78.10\%} & \textbf{+1.13\%}\\
\hline
\end{tabular}}
\end{small}
\end{center}
\caption{Scene text segmentation performances using synthetic data and/or the proposed datasets. The notation "$+$ Dataset" indicates that a fine--tune procedure has been carried out on "Dataset". The last column reports the relative F1 score increment, with and without fine--tuning, compared to the use of synthetic data only.}
\end{table*}

\begin{figure*}[!ht]
\begin{center}
\centerline{
\includegraphics[width=2.6cm,keepaspectratio]{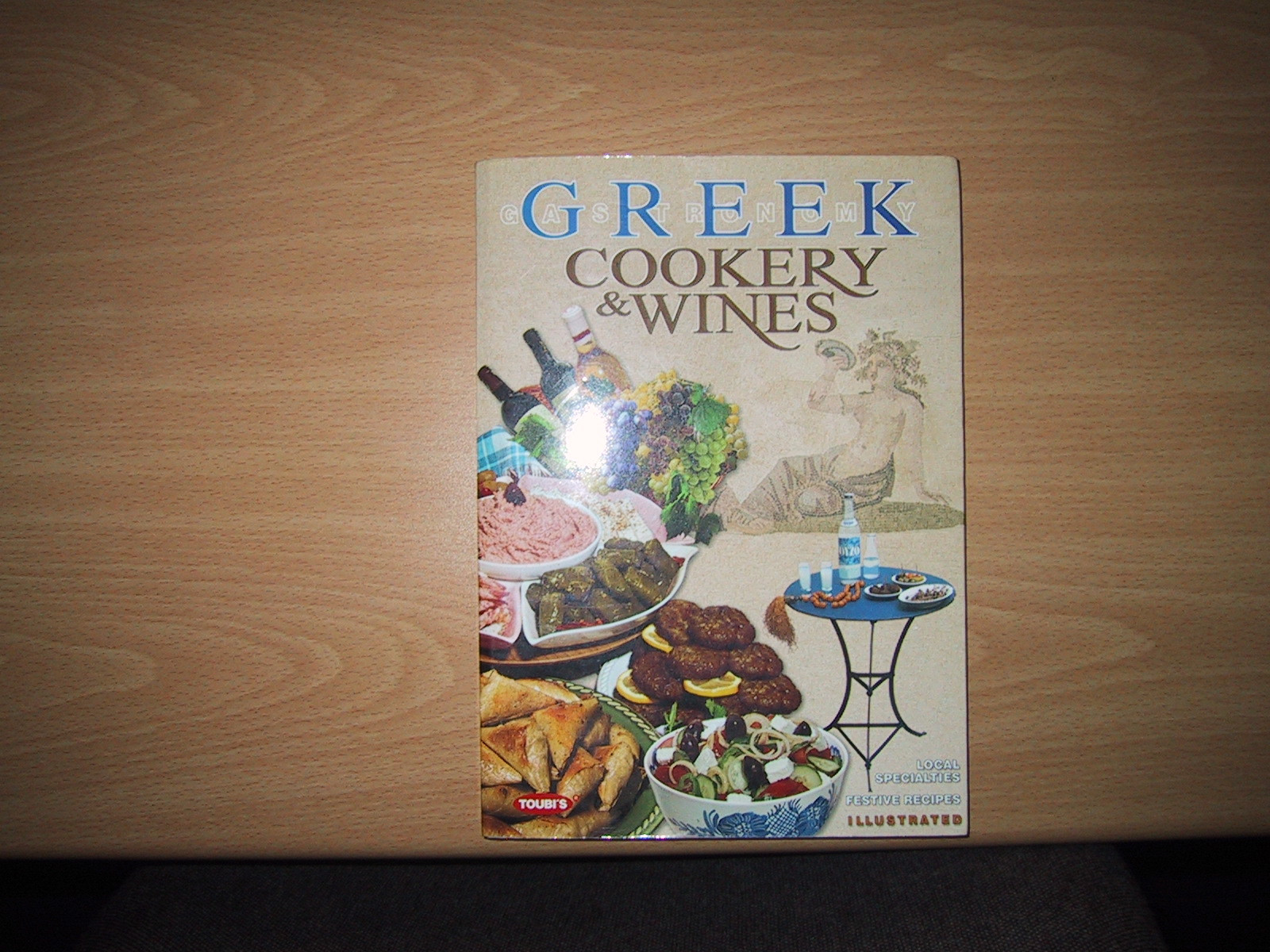}
\includegraphics[width=2.6cm,keepaspectratio]{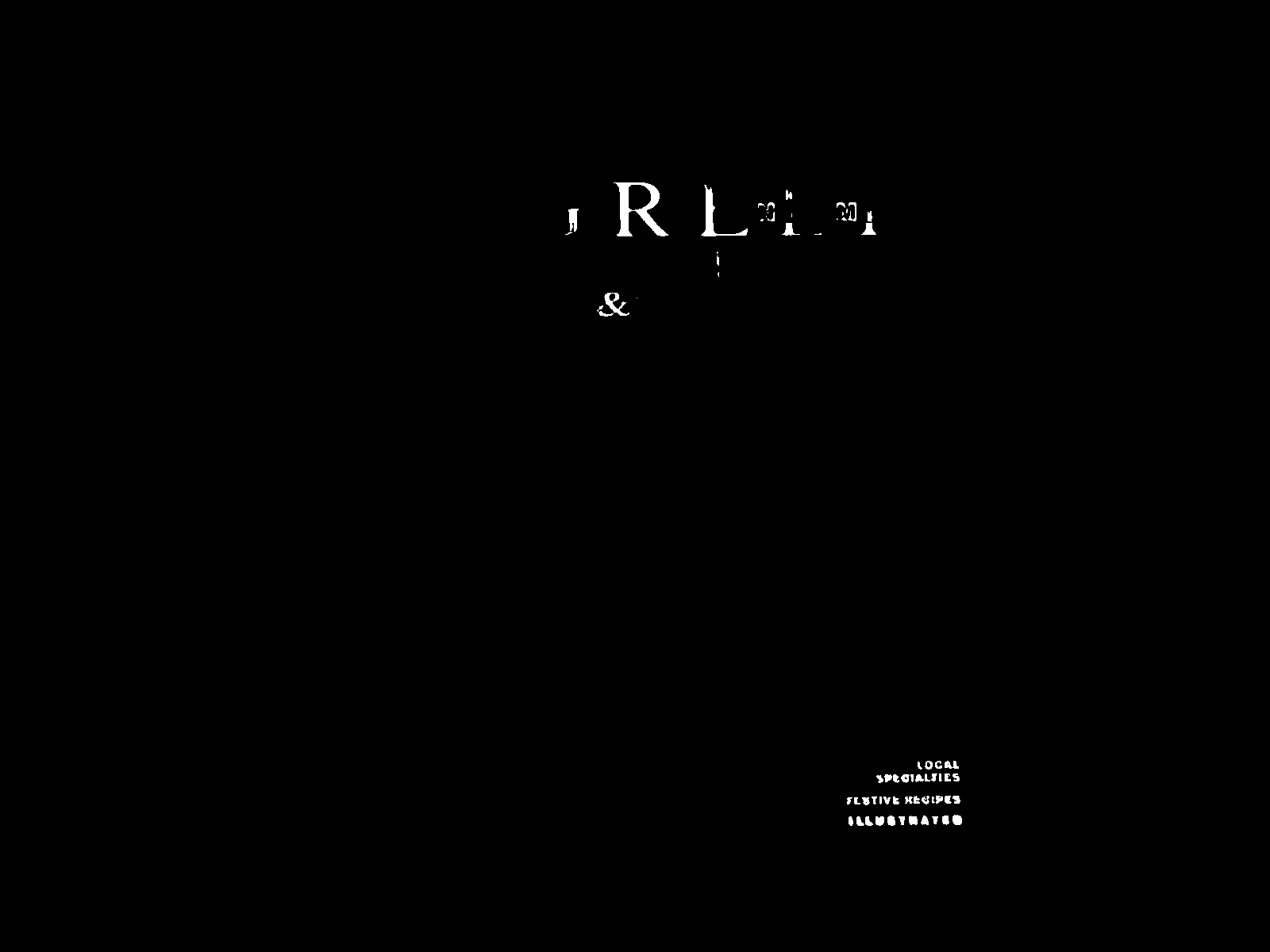}
\includegraphics[width=2.6cm,keepaspectratio]{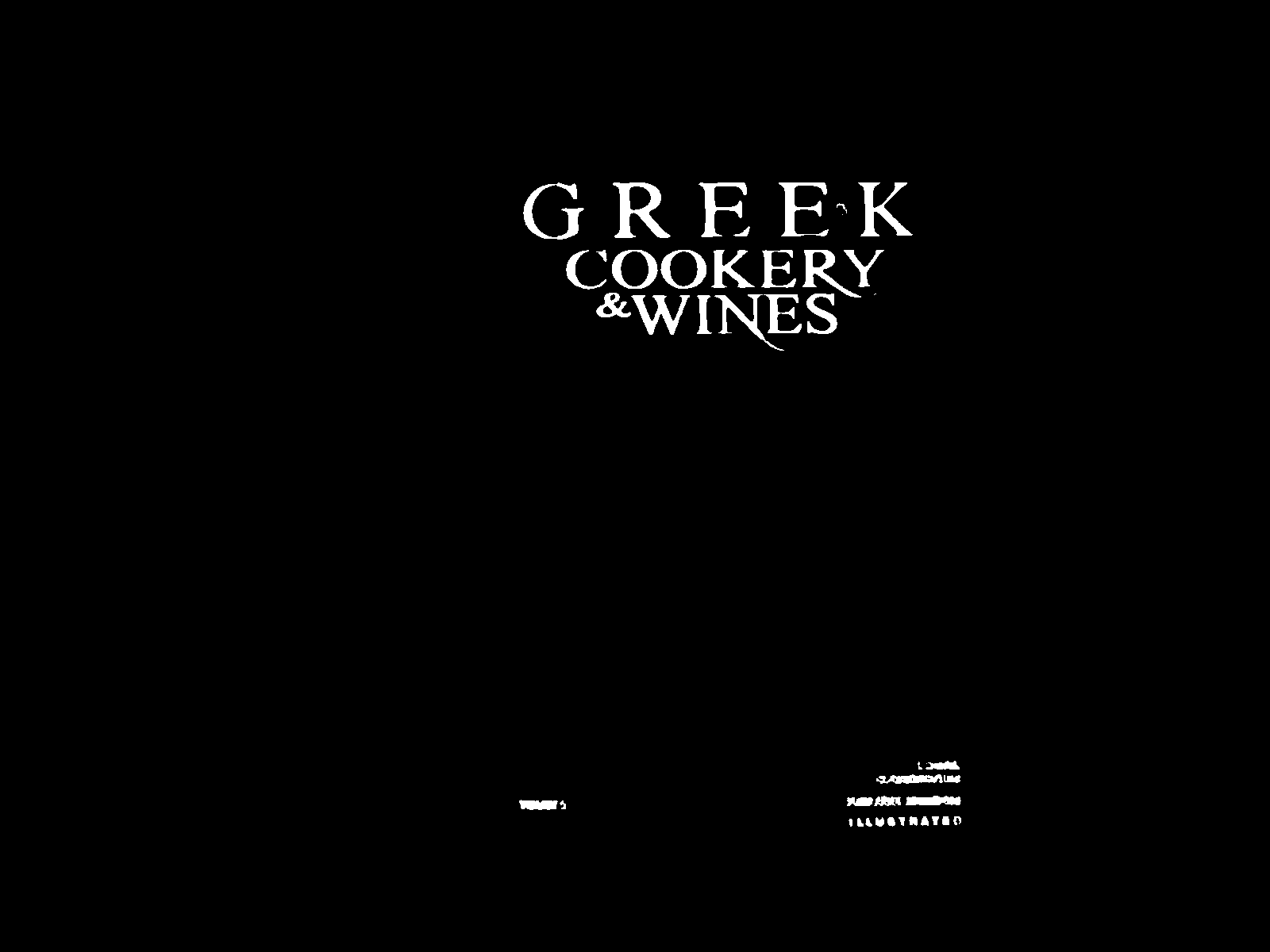}
\includegraphics[width=2.6cm,keepaspectratio]{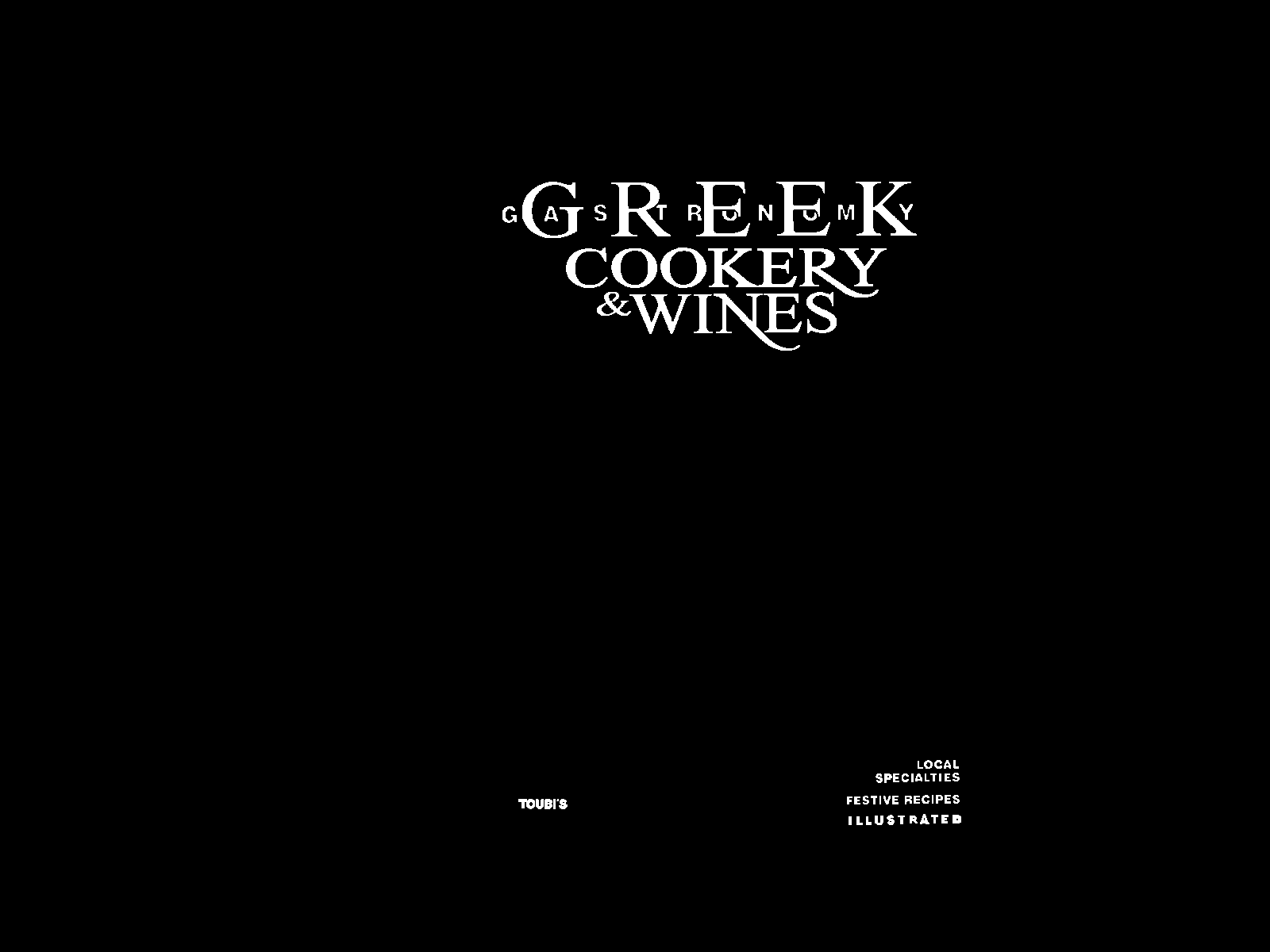}}
\vskip 0.08in
\centerline{
\includegraphics[width=2.6cm,keepaspectratio]{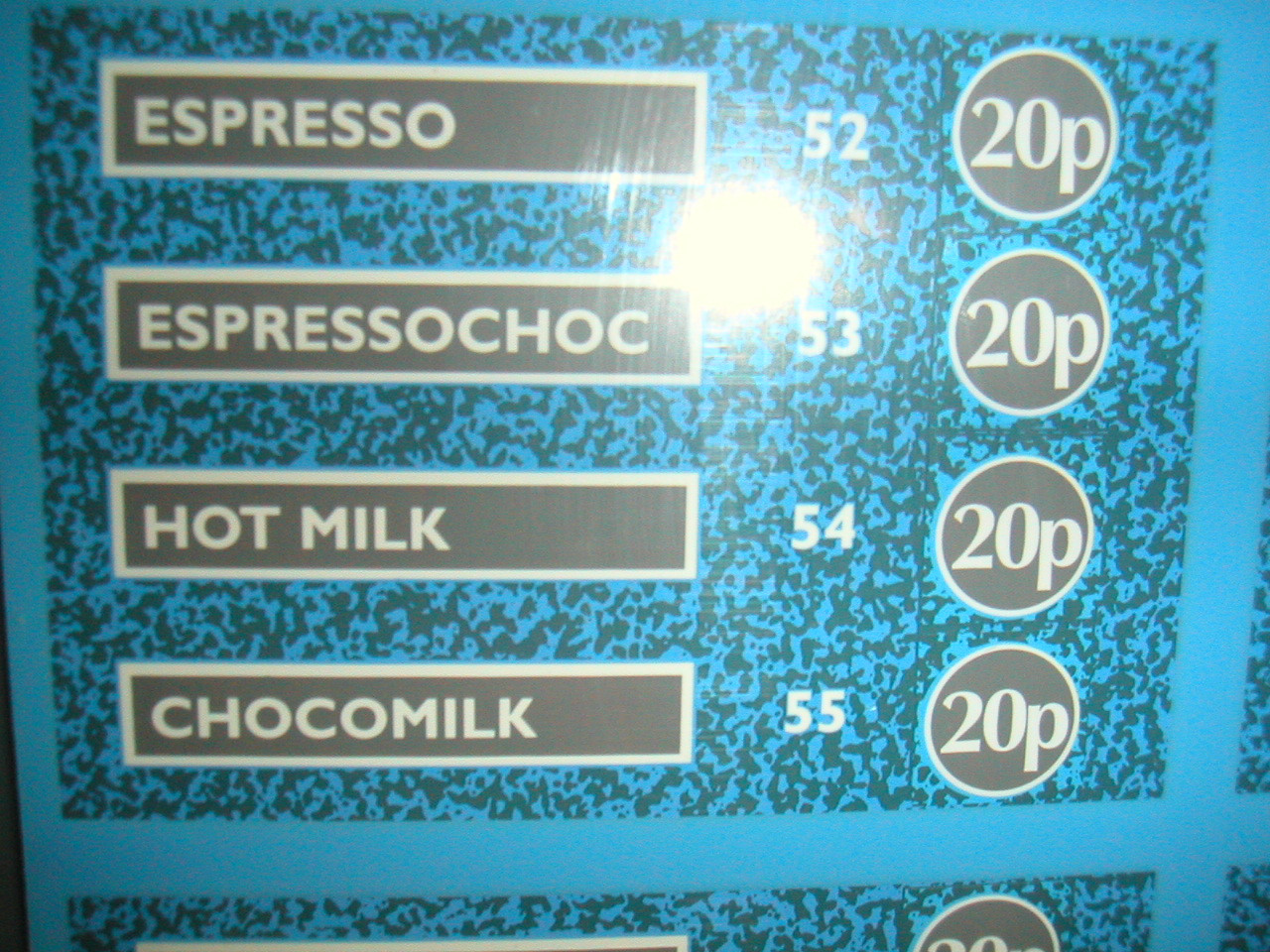}
\includegraphics[width=2.6cm,keepaspectratio]{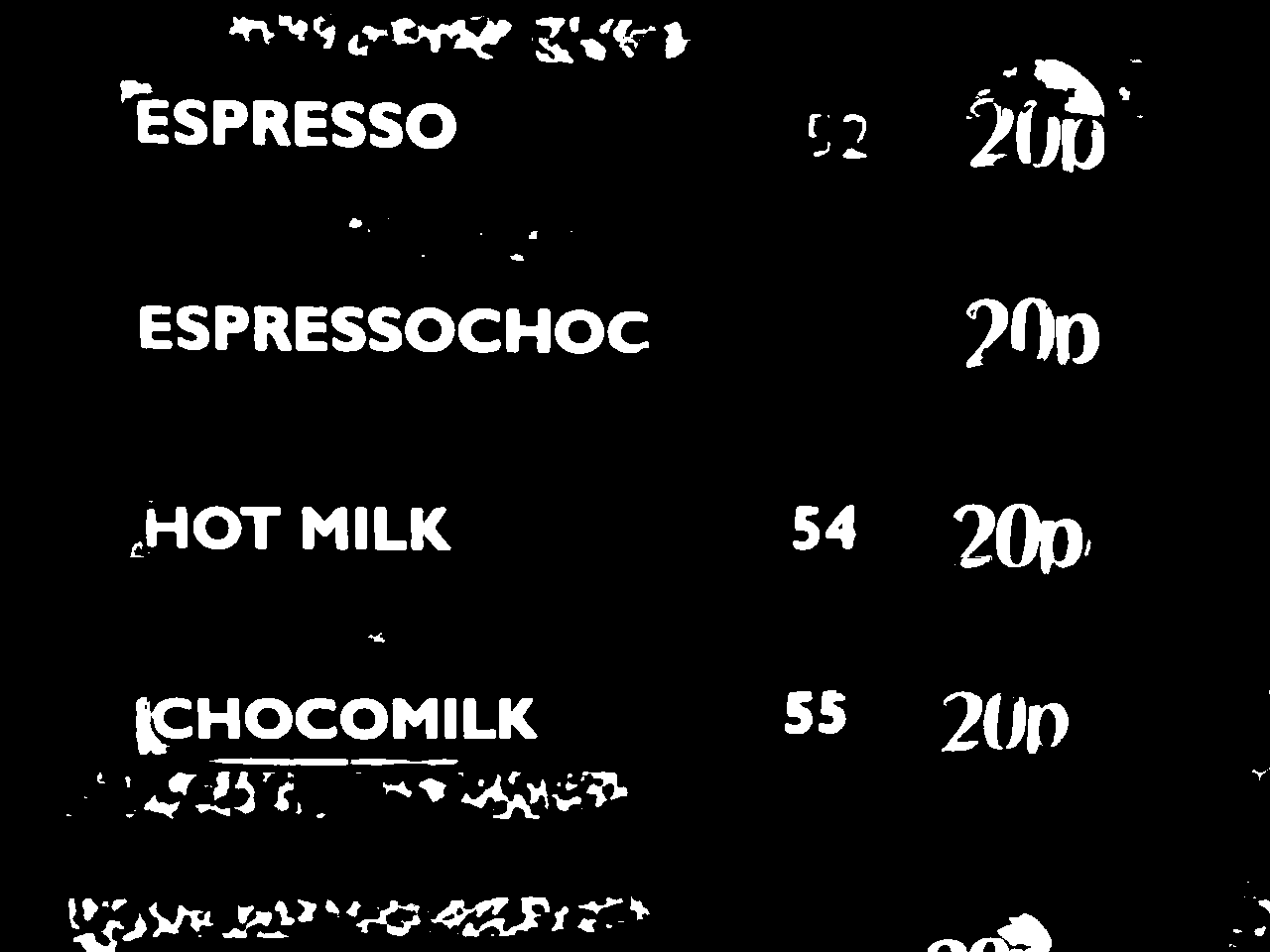}
\includegraphics[width=2.6cm,keepaspectratio]{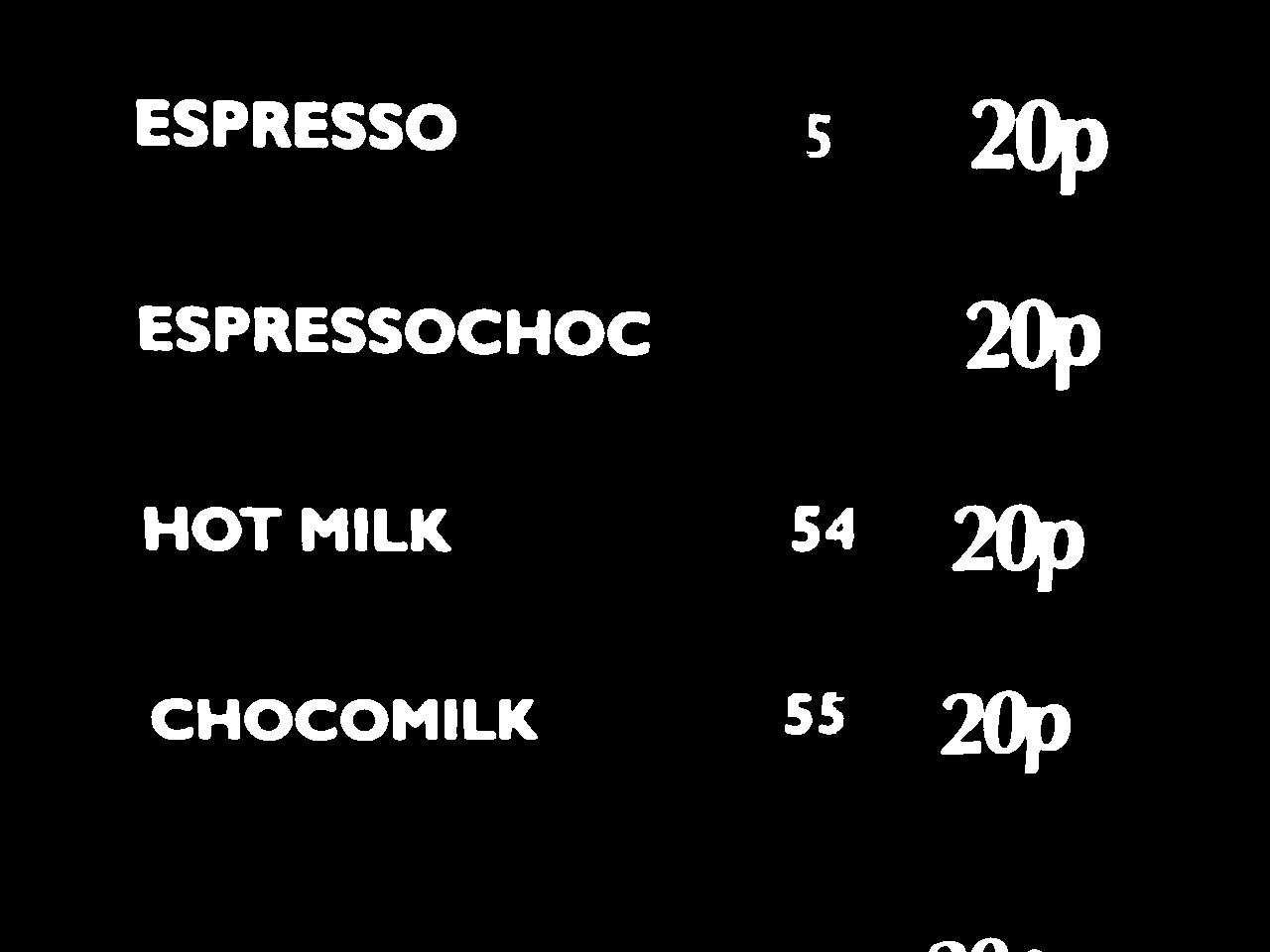}
\includegraphics[width=2.6cm,keepaspectratio]{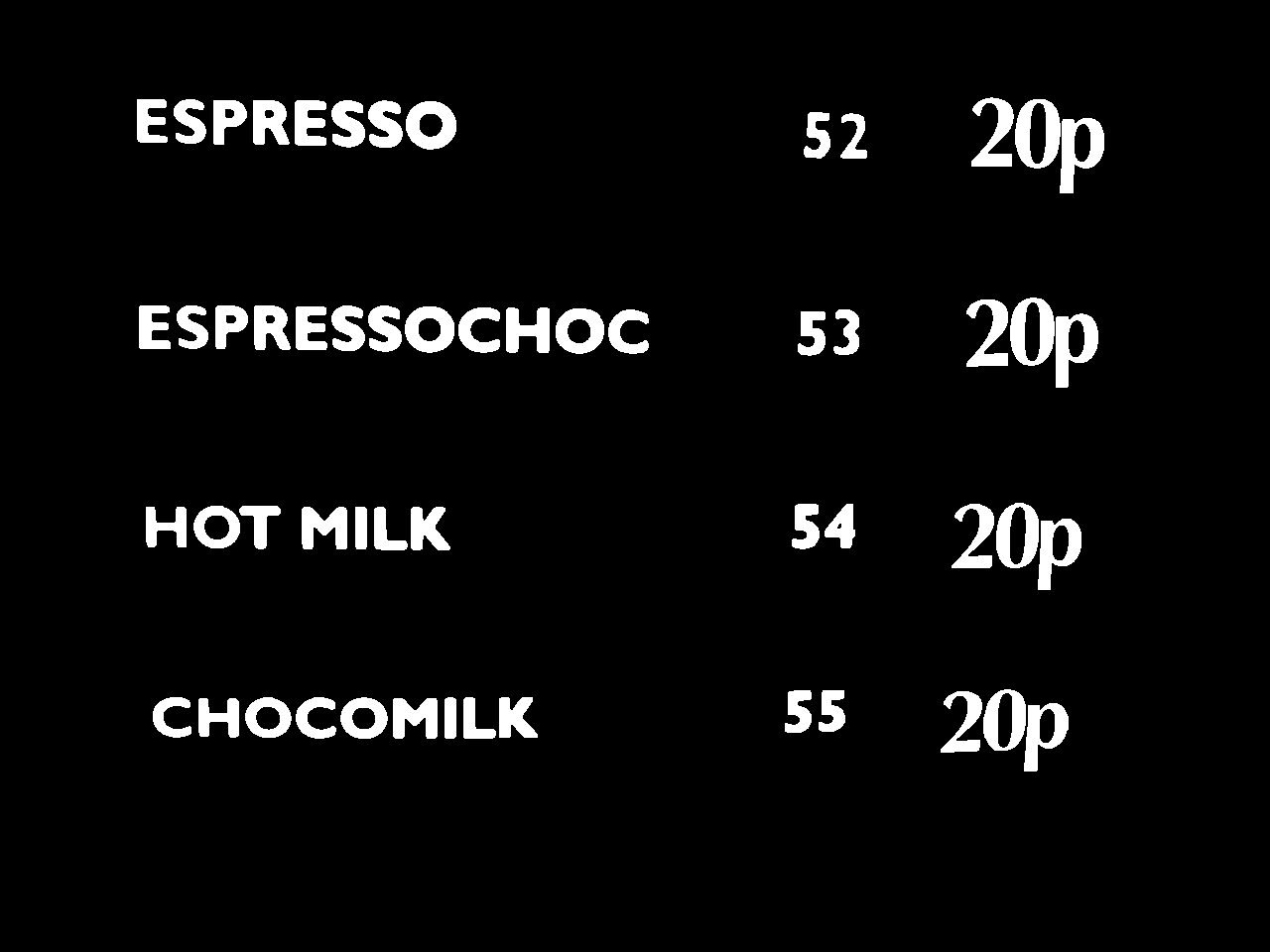}}
\vskip 0.08in
\centerline{
\includegraphics[width=2.6cm,keepaspectratio]{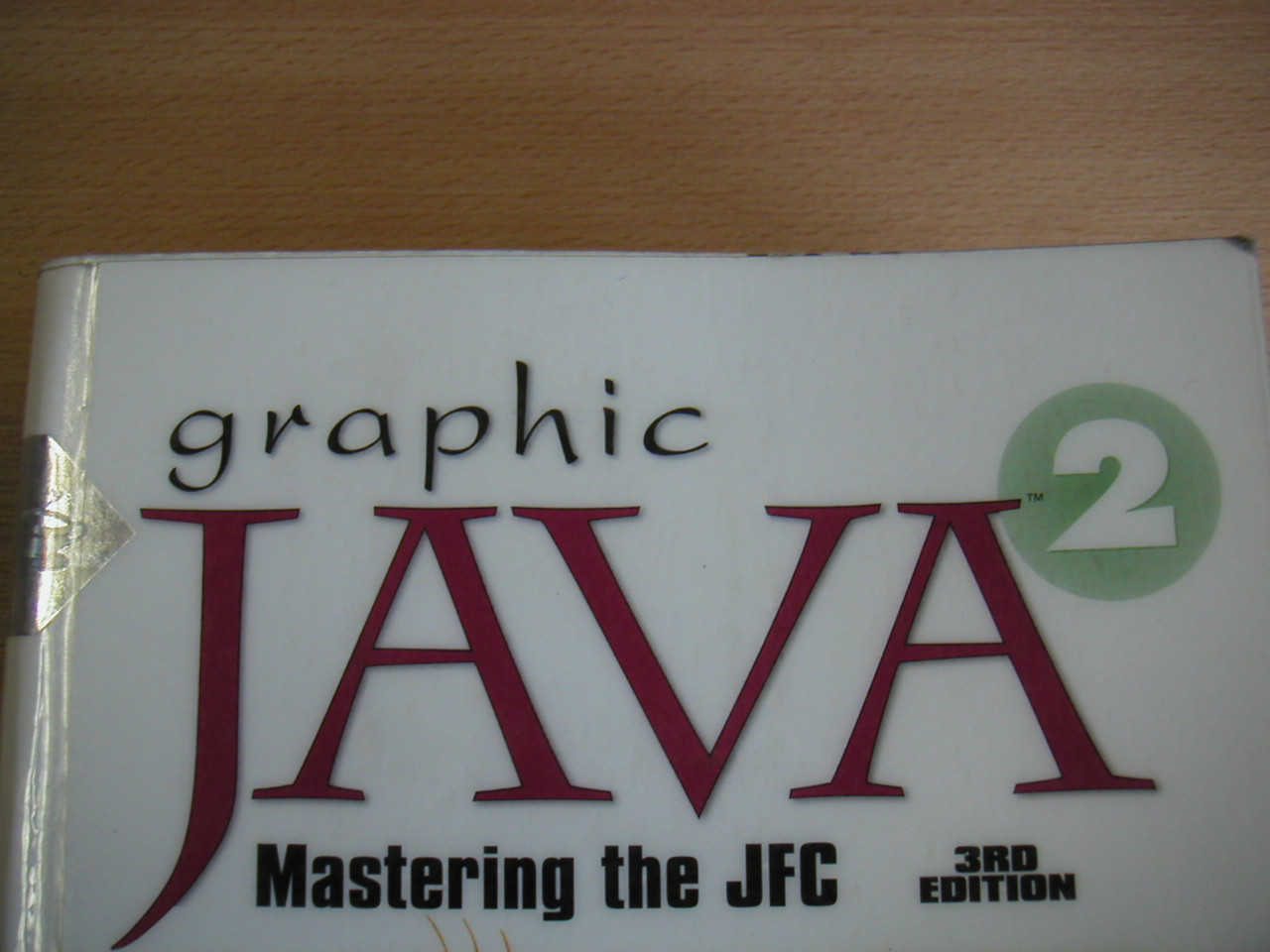}
\includegraphics[width=2.6cm,keepaspectratio]{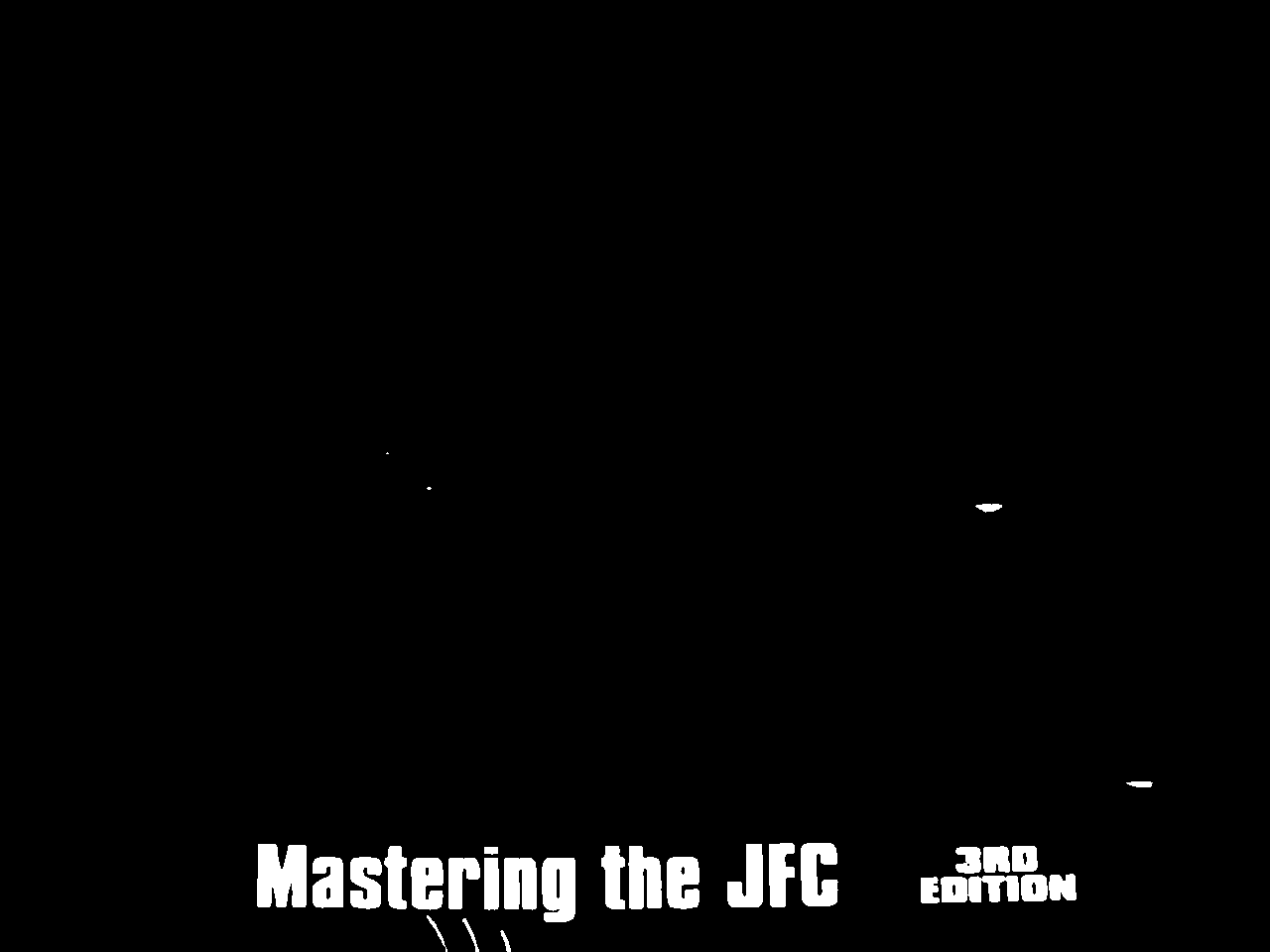}
\includegraphics[width=2.6cm,keepaspectratio]{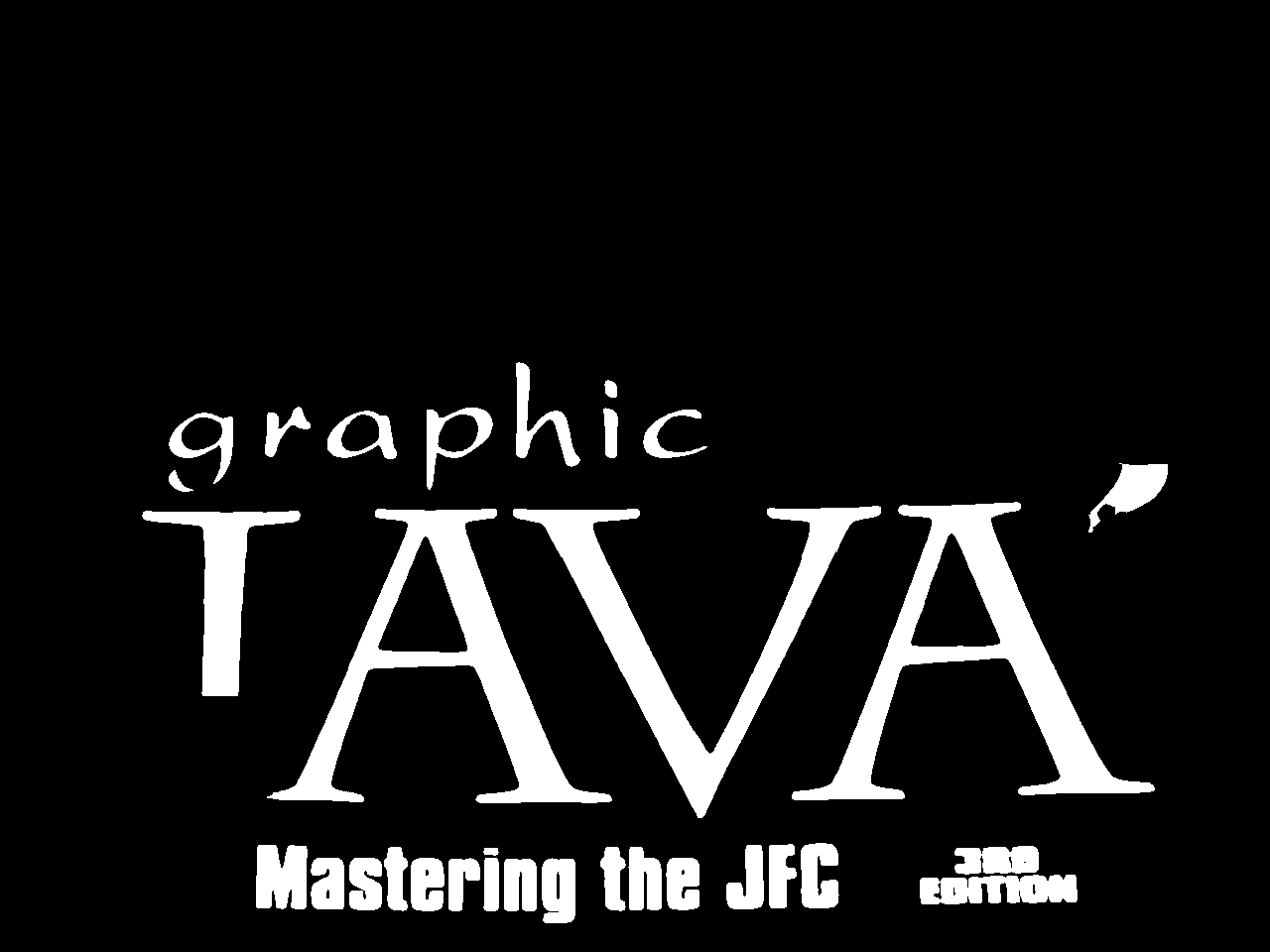}
\includegraphics[width=2.6cm,keepaspectratio]{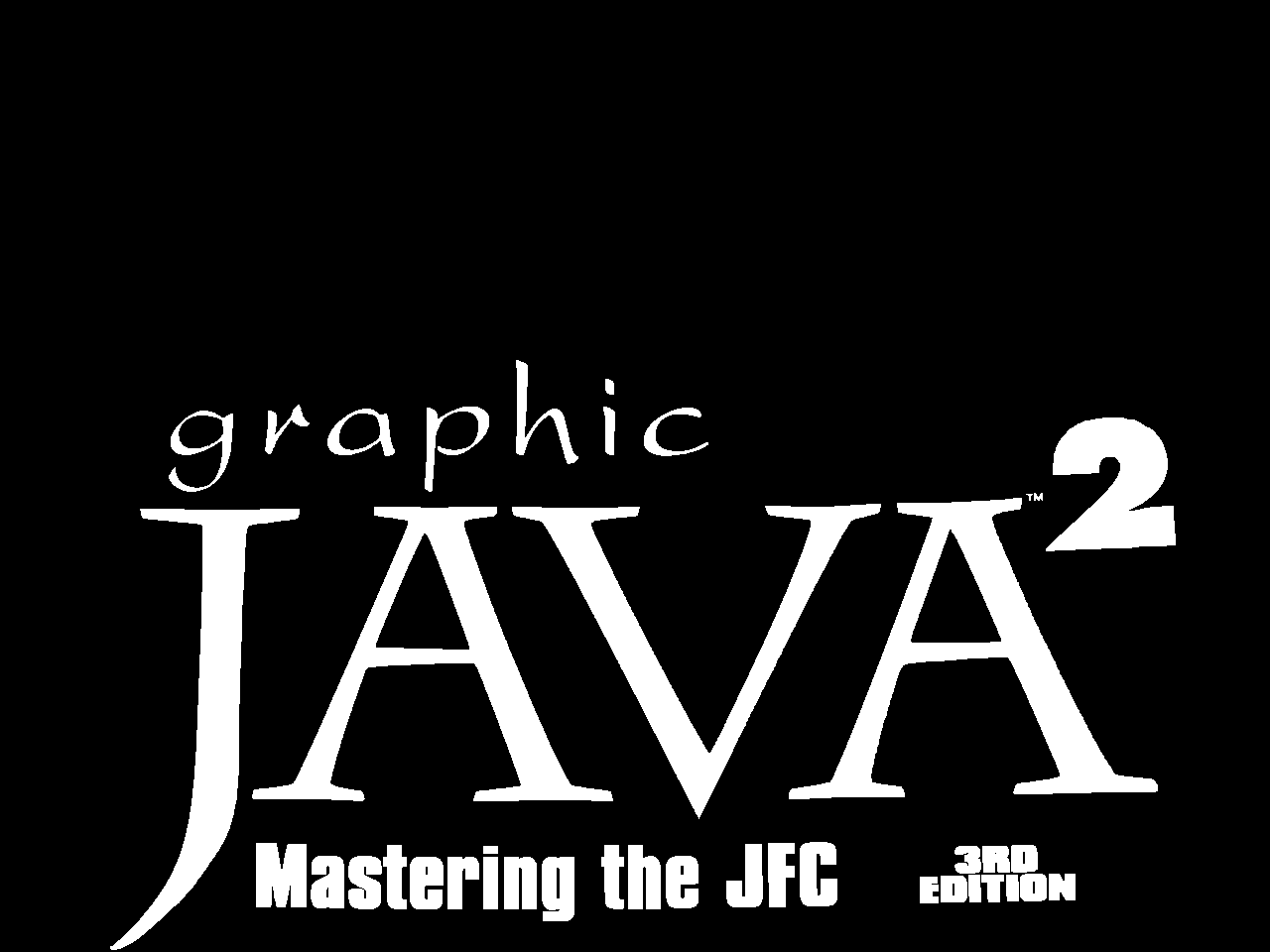}}
\vskip -0.06in
\centerline{
\subfloat[]{\includegraphics[width=2.6cm,keepaspectratio]{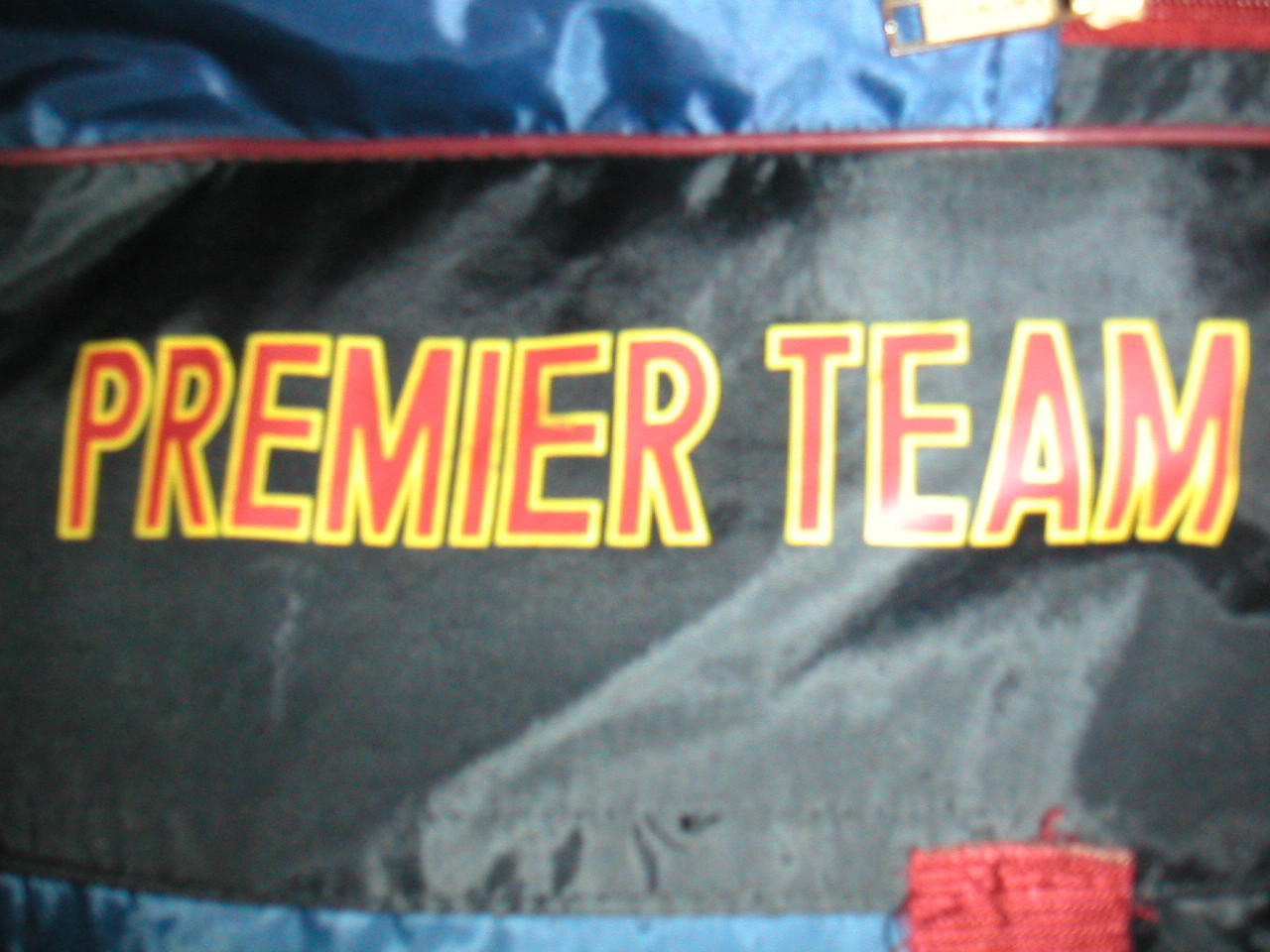}}\hskip 0.05in
\subfloat[]{\includegraphics[width=2.6cm,keepaspectratio]{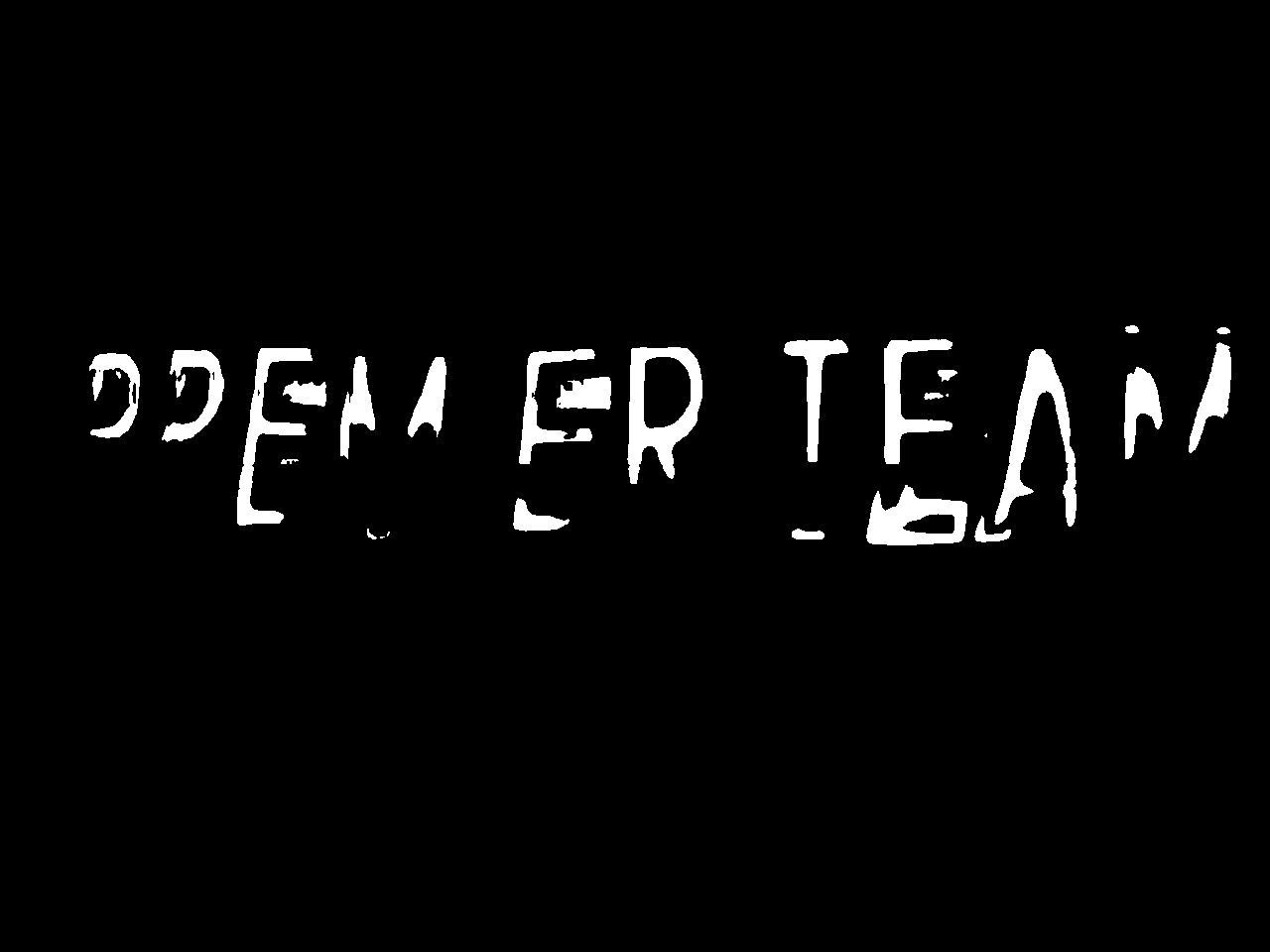}}\hskip 0.05in
\subfloat[]{\includegraphics[width=2.6cm,keepaspectratio]{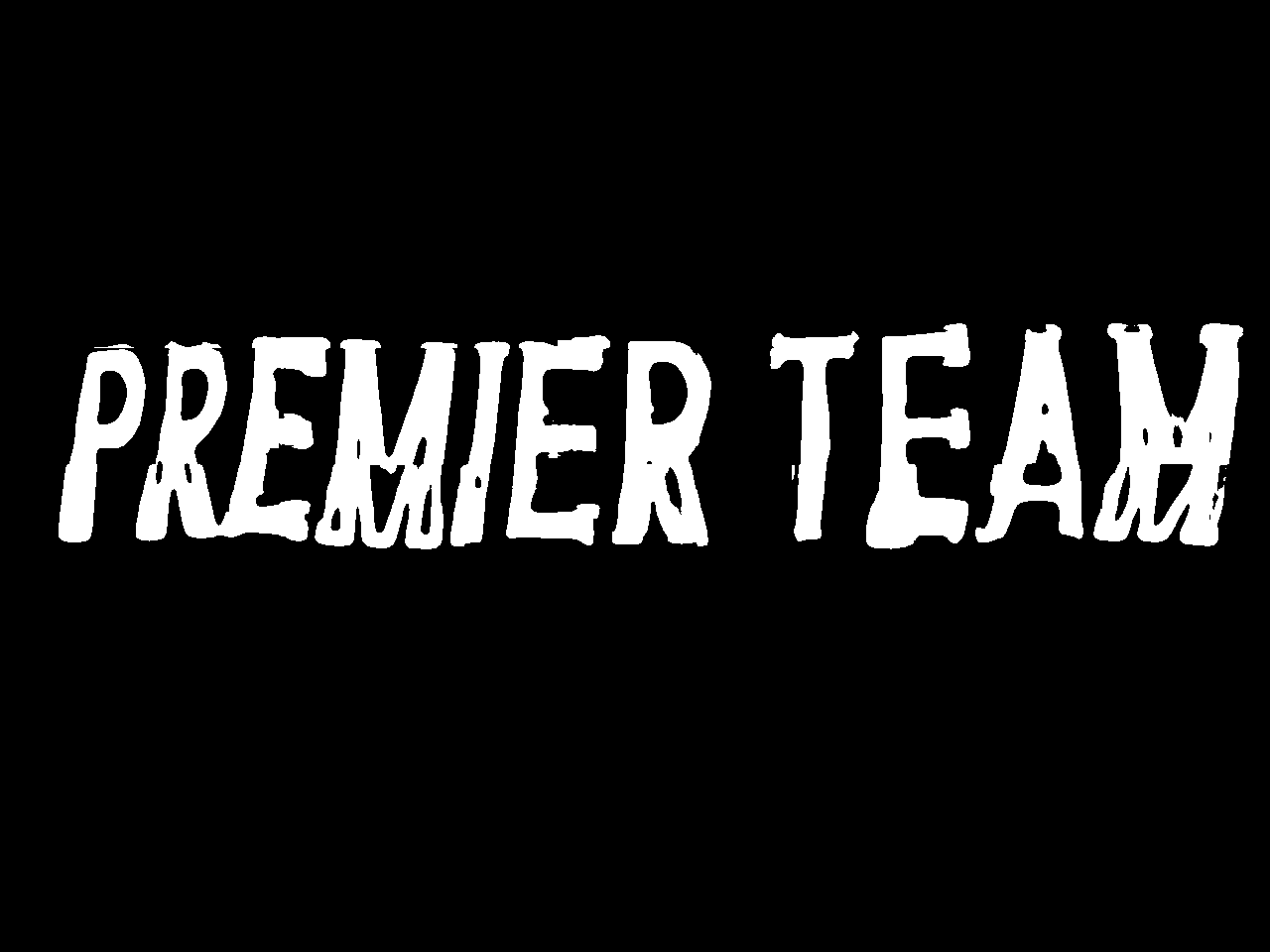}}\hskip 0.05in
\subfloat[]{\includegraphics[width=2.6cm,keepaspectratio]{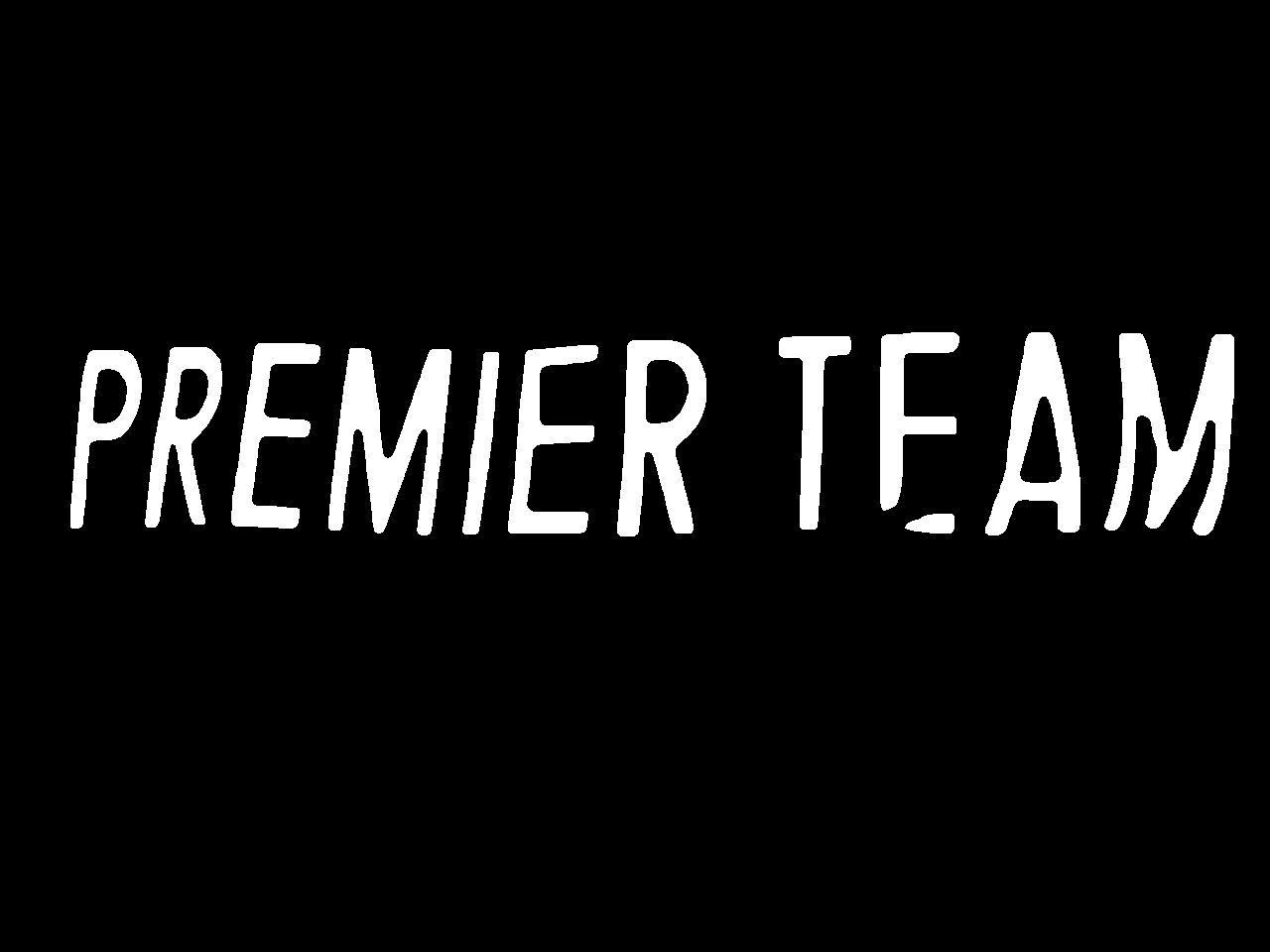}}}
\caption{Results on the ICDAR--2013 test set. In (a) the original image, in (b) and (c) the segmentation obtained with Synth and COCO\_TS + MLT\_S setups, respectively. The ground--truth supervision is reported in (d).}
\label{segmentation_results_ICDAR}
\end{center}
\begin{center}
\centerline{
\includegraphics[width=2.6cm,keepaspectratio]{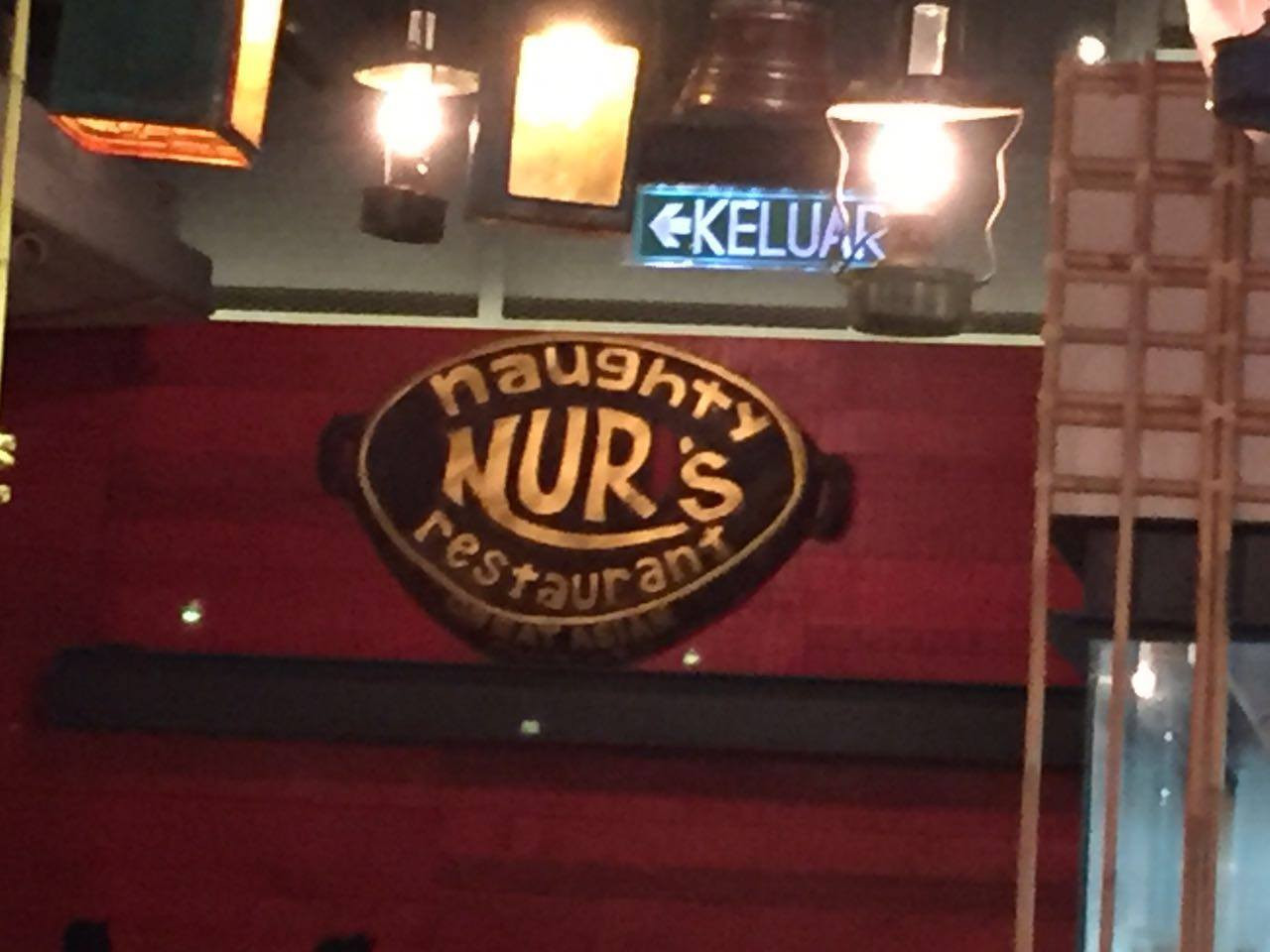}
\includegraphics[width=2.6cm,keepaspectratio]{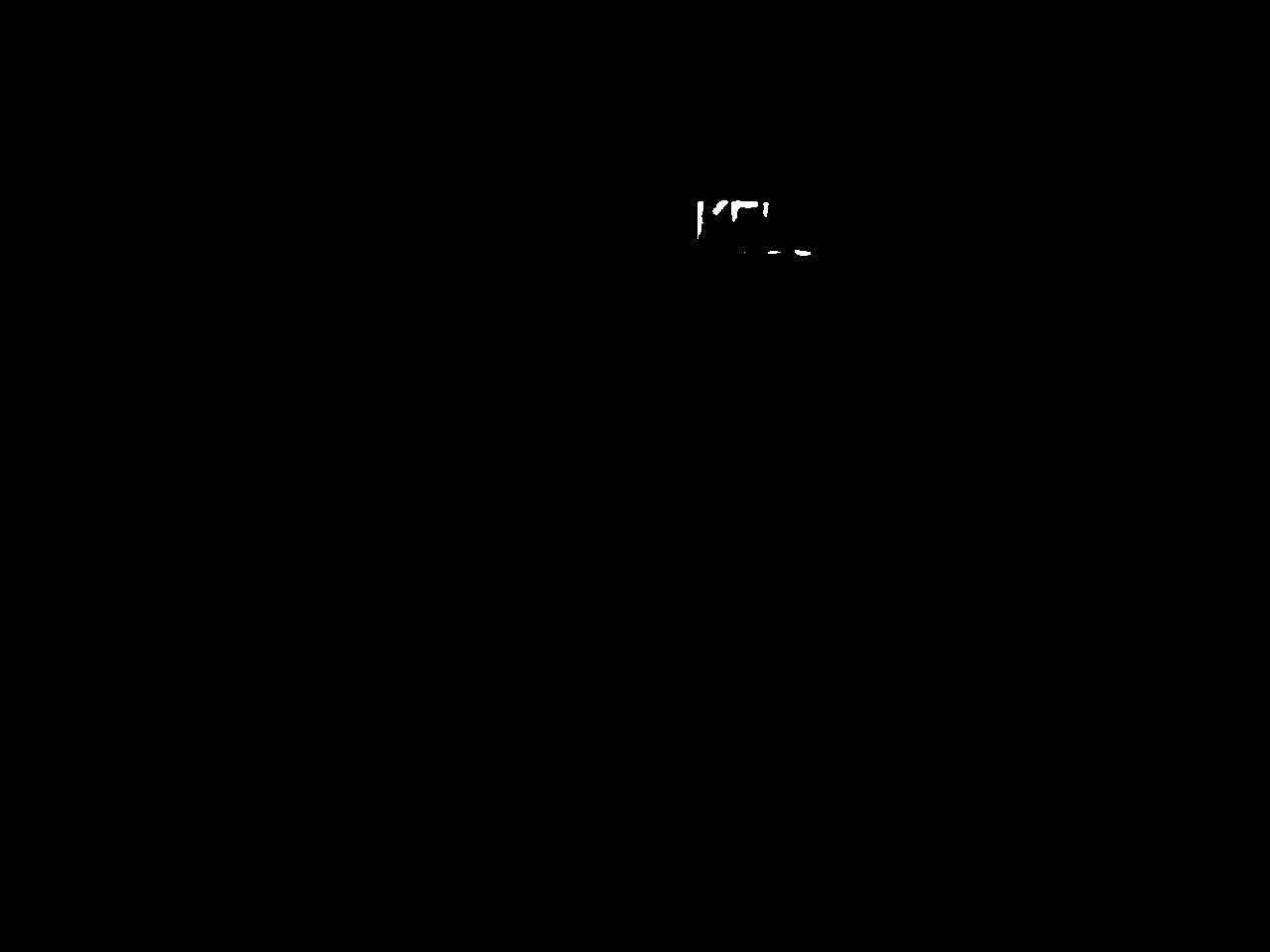}
\includegraphics[width=2.6cm,keepaspectratio]{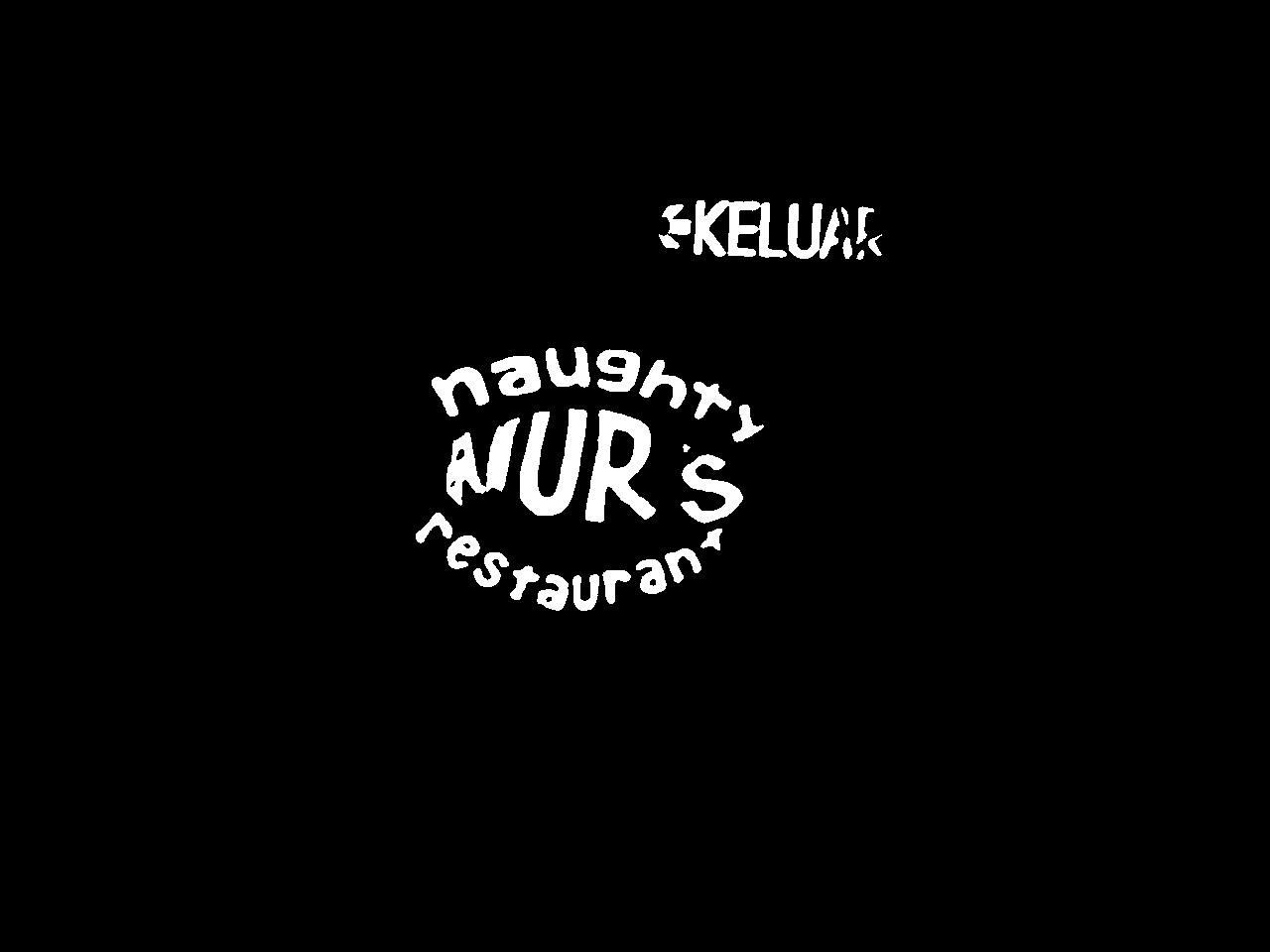}
\includegraphics[width=2.6cm,keepaspectratio]{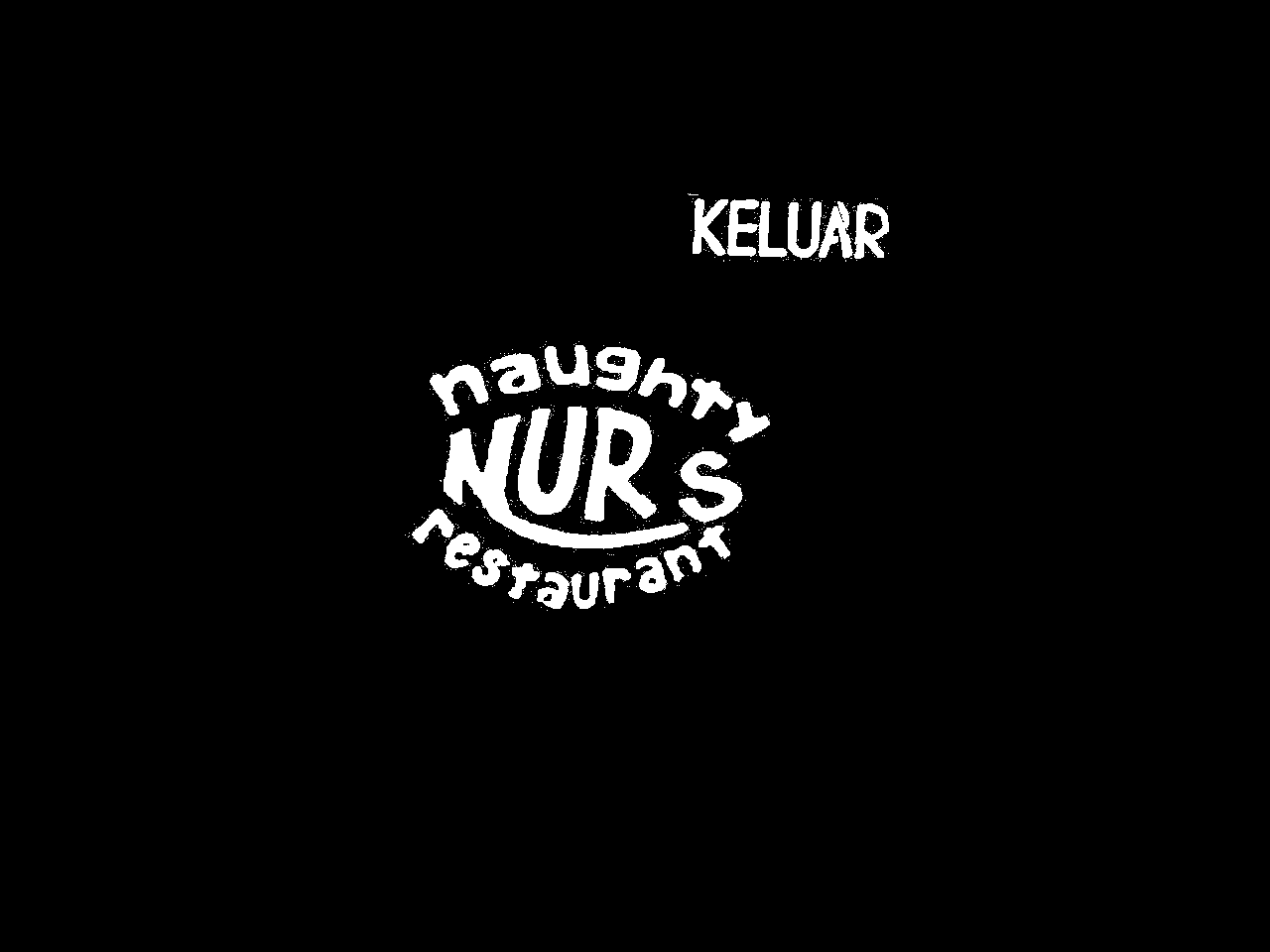}}
\vskip 0.08in
\centerline{
\includegraphics[width=2.6cm,keepaspectratio]{img_segmentation/total_text/src/img6.jpg}
\includegraphics[width=2.6cm,keepaspectratio]{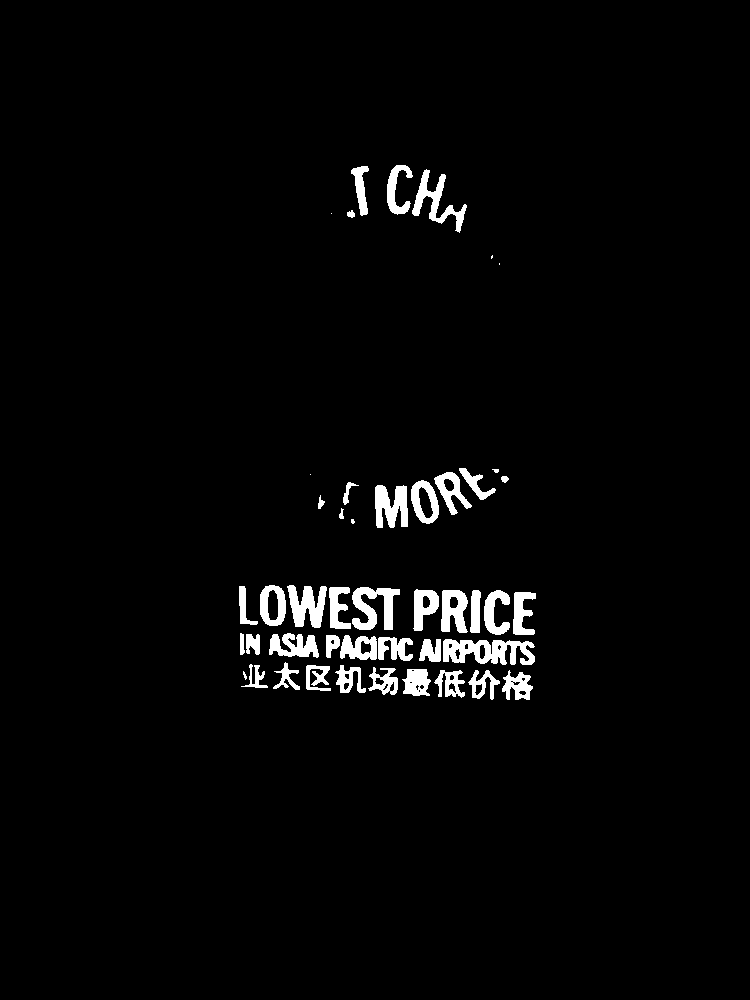}
\includegraphics[width=2.6cm,keepaspectratio]{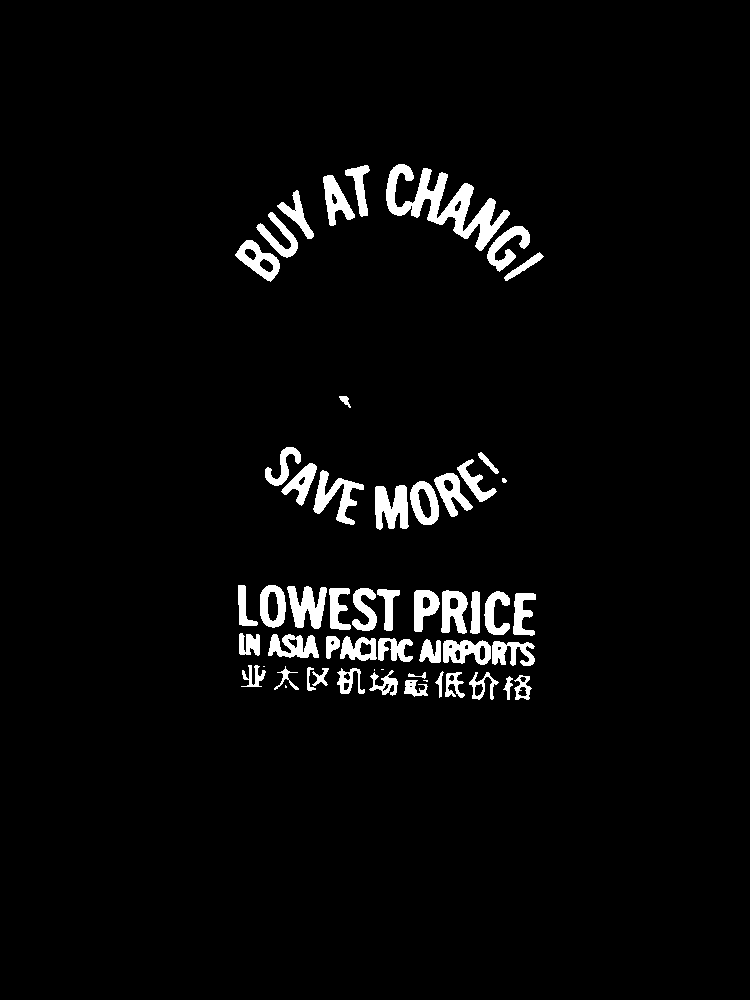}
\includegraphics[width=2.6cm,keepaspectratio]{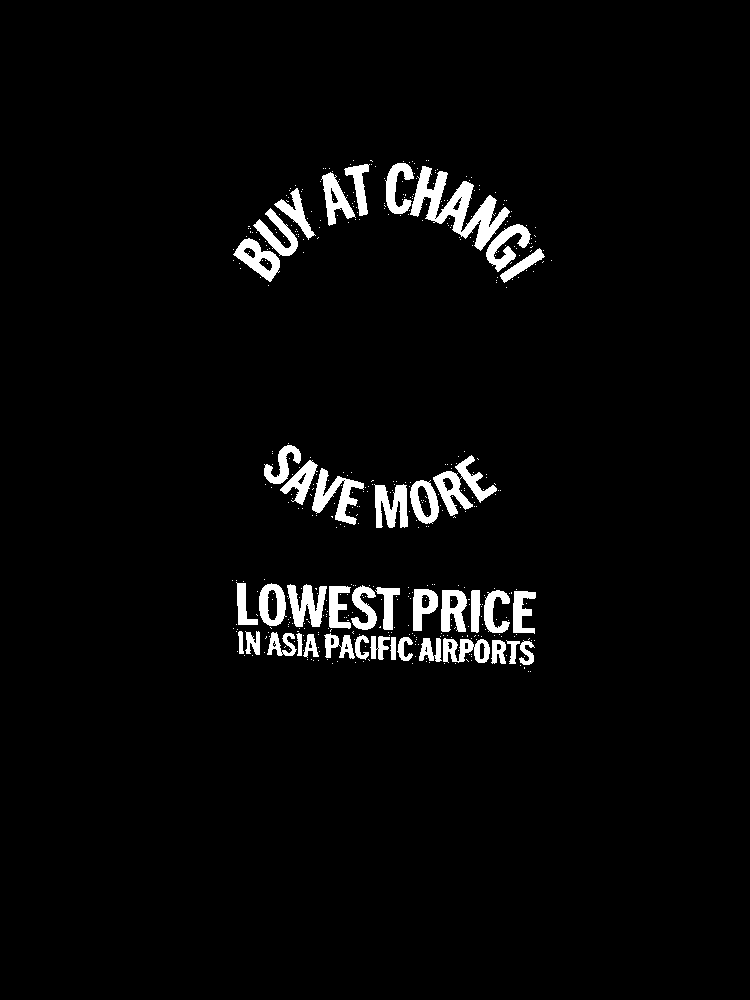}}
\vskip 0.08in
\centerline{
\includegraphics[width=2.6cm,keepaspectratio]{img_segmentation/total_text/src/img91.jpg}
\includegraphics[width=2.6cm,keepaspectratio]{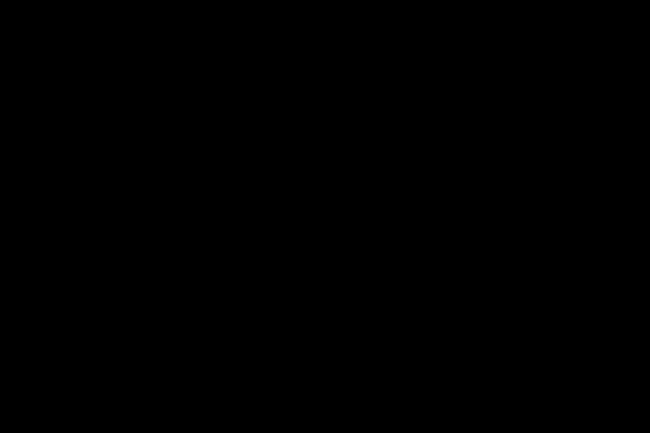}
\includegraphics[width=2.6cm,keepaspectratio]{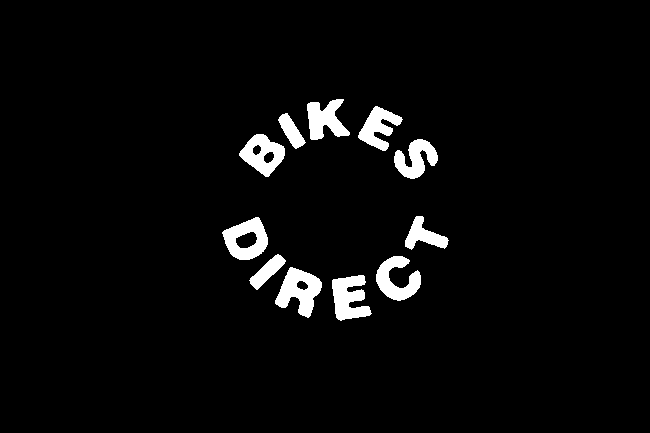}
\includegraphics[width=2.6cm,keepaspectratio]{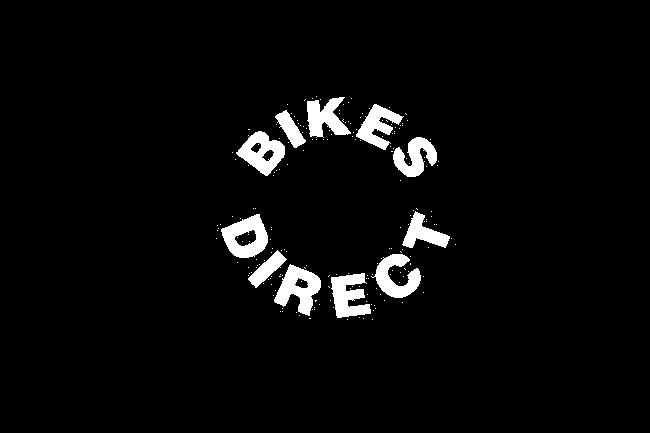}}
\vskip -0.06in
\centerline{
\subfloat[]{\includegraphics[width=2.6cm,keepaspectratio]{img_segmentation/total_text/src/img531.jpg}}\hskip 0.05in
\subfloat[]{\includegraphics[width=2.6cm,keepaspectratio]{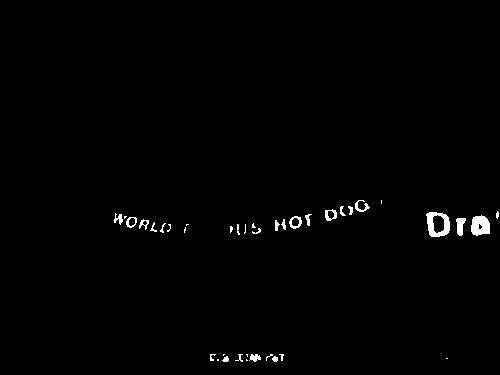}}\hskip 0.05in
\subfloat[]{\includegraphics[width=2.6cm,keepaspectratio]{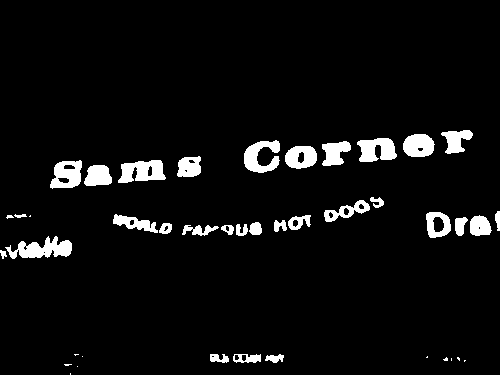}}\hskip 0.05in
\subfloat[]{\includegraphics[width=2.6cm,keepaspectratio]{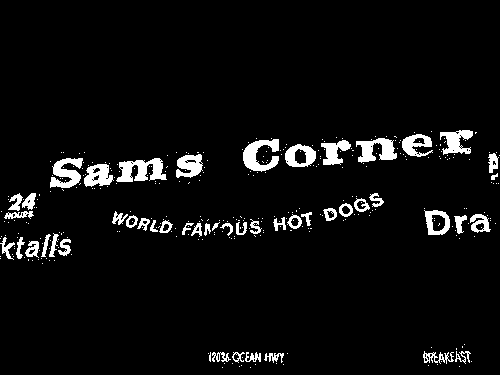}}}

\caption{Results on the Total--Text test set. In (a) the original image, in (b) and (c) the segmentation obtained with Synth and COCO\_TS + MLT\_S setups, respectively. The ground--truth supervision is reported in (d).}
\label{segmentation_results_TotalText}
\end{center}
\end{figure*}

\noindent It is worth noting that, training the network using both the datasets, COCO\_TS and MLT\_S, is more effective than using synthetic images. Specifically, employing together the proposed datasets, the F1 Score is improved of 17.82\% and 28.33\% on ICDAR--2013 and Total--Text, respectively. 
These results are quite surprising and prove that the proposed datasets substantially increase the network performance, reducing the domain--shift from synthetic to real images. If the network is fine--tuned on ICDAR--2013 and Total--Text, the relative difference between the use of synthetic images and the COCO\_TS dataset is reduced, but still remains significant. Specifically, the F1 Score is improved by 7.34\% on ICDAR--2013 and 1.13\% on Total--Text. 
\noindent 
Furthermore, it can be observed that the best result has been obtained using both, COCO\_TS and MLT\_S, which means that the two datasets are complementary and prove to be a valid alternative to synthetic data generation for scene text segmentation. Moreover, the use of real images increases the sample efficiency, allowing to substantially reduce the number of samples needed for training. In particular, the COCO\_TS and the MLT\_S together contain 21,586 samples that are less than 1/37 of the synthetic dataset cardinality.  
Some qualitative output results of the scene text segmentation network are shown in Figure \ref{segmentation_results_ICDAR} and Figure \ref{segmentation_results_TotalText}.

\subsection{SMANet evaluation}
\label{smanet_eval}
The results reported in the previous sections are obtained using the SMANet architecture, which was specifically designed to deal with peculiarities of scene text segmentation (i.e. written text at different scales and thin text). To evaluate the effectiveness of the modules added to the SMANet --- the attention module and the double decoder --- an ablation study was carried out. To perform this study we chose the best setup both on ICDAR--2013 and on Total--Text (pre--training on both COCO\_TS and MLT\_S and then fine--tuning alternatively on ICDAR--2013 and on Total--Text). A comparison of the results obtained with three different network architectures, PSP--Net (baseline), PSP--Net with double decoder and SMANet, on ICDAR--2013 and Total--Text are reported in Table \ref{smanet_res_icdar} and Table \ref{smanet_res_total_text}, respectively.

\begin{table*}[!ht]
\label{smanet_res}
\begin{center}
\begin{small}
\subfloat[\small{Scene text segmentation performances on ICDAR--2013 test set.} \label{smanet_res_icdar}]{%
\begin{tabular}{lcccc}
\toprule
 & Precision & Recall & F1 Score & \\
\midrule
PSP--Net & 78.50\%& 72.40\% & 75.32\% & --\\
PSP--Net with double decoder & 85.40\% & 82.00\% & 83.66\% & +8.34\% \\
\textbf{SMANet} & 87.30\% & 84.40\% & \textbf{85.82\%} & \textbf{+10.45\%} \\
\bottomrule
\end{tabular}}
\end{small}

\begin{small}
\subfloat[\small{Scene text segmentation performances on Total--Text test set.} \label{smanet_res_total_text}]{%
\begin{tabular}{lcccc}
\toprule
 & Precision & Recall & F1 Score & \\
\midrule
PSP--Net & 73.20\%& 53.20\% & 61.62\% & --\\
PSP--Net with double decoder & 85.90\% & 70.40\% & 77.39\% & +15.77\% \\
\textbf{SMANet} & 86.00\% & 71.50\% & \textbf{78.10\%} & \textbf{+16.48\%} \\
\bottomrule
\end{tabular}}
\end{small}
\end{center}
\caption{The PSP--Nets and the SMANet are pre--trained on both COCO\_TS and MLT\_S datasets and fine--tuned on ICDAR--2013 (\ref{smanet_res_icdar}) and on Total--Text training sets (\ref{smanet_res_total_text}). The last column of each table reports the relative F1 score increment compared with the PSP--Net used as baseline.}
\end{table*}

\noindent
The results show that the addition of the \say{double decoder} significantly increases the performances compared with the baseline given by the PSP--Net (8.34\% in F1 score on ICDAR--2013 and 15.77\% on Total--Text). The decoder improves the network capability of recovering fine details, which is important in natural images where text can exhibit a variety of different dimensions. Employing the multi--scale attention module the results are further improved (2.11\% F1 score on ICADAR-2013 and 0.71\% on Total--Text). As a concluding remark, we can assert that, even if comparatively assessing the SMANet performance for text segmentation is out of the scope of this paper, nonetheless, all its constitutive modules are fundamental for the whole architecture to be tailored to the considered task.

\section{Conclusions} \label{Conclusions}
This paper presents the generation of two new pixel--level annotated datasets and propose a tailored scene text segmentation network. A weakly supervised learning approach is employed to automatically convert the bounding--box annotations of a real dataset to pixel--level supervisions. The COCO\_TS and the MLT\_S datasets, which contain, respectively, the segmentation ground--truth of a subset of COCO--Text and MLT, are generated. The experiments, demonstrate the effectiveness of the proposed datasets showing a very significant improvement in generalization on both the ICDAR--2013 and Total--Text datasets, although with only a fraction of the samples that would be required using only synthetic data.
To foster further research on scene text segmentation, the COCO\_TS and the MLT\_S datasets have been released. Moreover, we employed a specifically designed architecture, the SMANet, that proved to be particularly effective for scene text segmentation.
It is a matter of future work to employ a region proposal network (f.i. Yolo \cite{yolov3}) to first automatically extract the bounding--box around the text region and then, using the proposed method, generate pixel--level supervisions of text in natural images.


\end{document}